\pdfoutput=1

\documentclass[11pt]{article}

\usepackage[preprint]{coling}

\usepackage{times}
\usepackage{latexsym}

\usepackage[T1]{fontenc}

\usepackage[utf8]{inputenc}

\usepackage{microtype}

\usepackage{inconsolata}

\usepackage{graphicx}

\usepackage{amsmath}
\usepackage{amssymb}
\usepackage{amsfonts}
\usepackage{amsthm}
\usepackage{graphicx}
\usepackage{subfig}
\usepackage{fancyvrb}
\usepackage{listings}

\usepackage{color}
\definecolor{Red}{RGB}{237,125,49}
\definecolor{Green}{RGB}{0,176,80}
\definecolor{Blue}{RGB}{0,176,240}

\title{Language Models Encode the Value of Numbers Linearly}

\author{Fangwei Zhu, Damai Dai, Zhifang Sui \\
  National Key Laboratory for Multimedia Information Processing, Peking University \\
  \texttt{zhufangwei2022@stu.pku.edu.cn} \\
  \texttt{\{daidamai, szf\}@pku.edu.cn} \\
}

\begin{document}
\maketitle
\begin{abstract}
Large language models (LLMs) have exhibited impressive competence in various tasks, but their internal mechanisms on mathematical problems are still under-explored.
In this paper, we study a fundamental question: how language models encode the value of numbers, a basic element in math.
To study the question, we construct a synthetic dataset comprising addition problems and utilize linear probes to read out input numbers from the hidden states.
Experimental results support the existence of encoded number values in LLMs on different layers, and these values can be extracted via linear probes.
Further experiments show that LLMs store their calculation results in a similar manner, and we can intervene the output via simple vector additions, proving the causal connection between encoded numbers and language model outputs.
Our research provides evidence that LLMs encode the value of numbers linearly, offering insights for better exploring, designing, and utilizing numeric information in LLMs. 
The code and data are available at \url{https://github.com/solitaryzero/NumProbe}.
\end{abstract}

\section{Introduction}

Large language models (LLMs) have demonstrated excellent ability in various scenarios like question answering~\cite{zhao2023verify, li2023chain}, instruction following~\cite{brown2020language, ouyang2022training, taori2023stanford}, and code generation~\cite{chen2021evaluating, nijkamp2022codegen, li2023starcoder}.
Solving mathematical problems is generally viewed to be more difficult~\cite{yu2023metamath}, and language models even struggle to solve simple arithmetic problems~\cite{dziri2024faith}.

Numbers are fundamental elements in math.
In order to accurately answer mathematical problems, LLMs should be able to precisely encode value of numbers in the input text.
Currently, the way how LLMs process numbers is still not fully explored.
While previous studies~\cite{stolfo2023mechanistic, hanna2024does} have explored the inner mechanisms of language models on mathematical problems, they focus on small numbers or modular arithmetic~\cite{engels2024not, zhong2024clock}, and how LLMs utilize numbers in a larger, unconstrained range remains largely unknown.

\begin{figure}[t]
    \centering
    \includegraphics[width=0.45\textwidth]{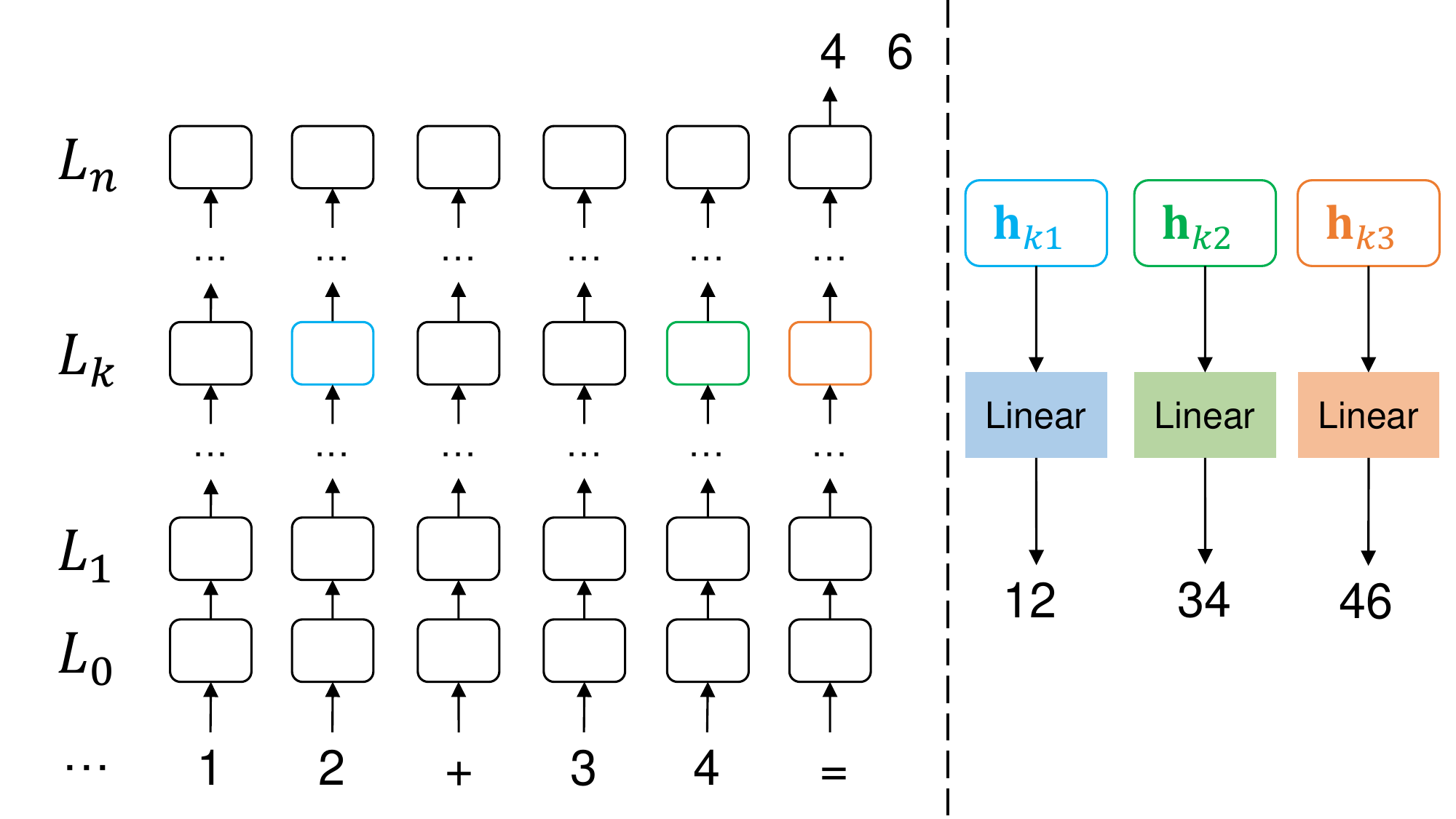}
    \caption{Encoded number values in the hidden state of language models. We find that both the value of input numbers (\textcolor{Blue}{blue} and \textcolor{Green}{green}) and calculation results (\textcolor{Red}{red}) can be read out from the hidden state of language models via linear probes.}
    \label{fig:datapath}
\end{figure}

In this paper, we explore the question whether and how LLMs encode the value of numbers through extracting numerical information from their internal representations.
To be specific, we construct a synthetic dataset comprising simple addition questions, and train linear probes~\cite{nanda2023emergent, gurnee2023language} on the hidden states of LLMs to predict the number values provided in the input text.
Experimental results on the dataset demonstrate that the value of input numbers can be probed from the hidden states of language models from early layers, as illustrated in Figure \ref{fig:datapath}.
Both input values and calculation results can be read out, and encoded values can be found at different token positions.
These results support that language models do encode numerical information, and possibly in a linear manner.

To further verify the fact that the encoded number values are utilized by language models, we study the causal connection between numeric information and model outputs.
To be specific, we discover that we can influence the calculation result of language models by performing interventions like activation patching or adding linear vectors.

The above discoveries may reveal future directions for utilizing the encoded numerical information, for example, specialized encoding systems and error mitigation modules.

To sum up, our contributions can be listed as:
(1) We study the question of whether language models are able to encode the value of numbers in the input text and construct a synthetic dataset to analyze the language models.
(2) We discover that language models encode the value of numbers linearly by utilizing linear probes to probe encoded number values in the hidden states of language models.
(3) We further prove that language models utilize the encoded numerical information by revealing the causal connection between encoded number values and the final output of language models.

\section{Probing Numbers in Language Models}
\subsection{The Goal of Probing}
Given that there is a number $x$ in the input text $t$, we assume that a language model $LM$ can encode the number in its hidden state $\mathbf{h}_i \in \mathbb{R}^{d_{model}}$ of a specific layer $i$, where $d_{model}$ is the hidden dimension. 
We denote the mapping as:
\begin{equation}
    \mathbf{h}_i = f_{i}(x, t-x)
\end{equation}
where $f_{i}$ refers to the encoding process on layer $i$, and $t-x$ refers to the tokens in $t$ excluding $x$.

If the mapping function $f$ is a bijective function, there will exist an inverse function $f^{-1}_{i}$ that reconstructs the original number $x$ from the hidden state $\mathbf{h}_i$.
For each layer $i$, we aim to find a optimal predictor $\mathcal{P}_{i}^{*}$ that imitates $f^{-1}_{i}$, whose prediction best fits the original number $x$:
\begin{equation}
    \mathcal{P}_{i}^{*} = \mathop{\arg\min}\limits_{\mathcal{P}_{i}} |x-\mathcal{P}_{i}(\mathbf{h}_i)| 
\end{equation}
Considering the numerical stability, we probe the logarithmic value $\log_{2}(x)$ instead of the original number $x$ in all our experiments.

We can assess the existence of encoded number values by observing how much the probing result $\mathcal{P}_{i}^{*}((\mathbf{h}_i))$ resembles the original number $x$.

\subsection{Dataset Construction}
\label{sec:dataset}
To investigate whether LLMs encode numbers, we construct a synthetic dataset containing different magnitudes of numbers.
The dataset contains numbers ranging from 2 digits to 10 digits, with each digit corresponding to 1000 entries\footnote{See Appendix \ref{sec:appendix_dataset} for more details.}.
We split the dataset into training, validation, and test sets at a ratio of 80\%/10\%/10\%.

To observe how LLMs encode and utilize numbers, we adopt addition problems as our prompt\footnote{The experimental results on subtraction problems are similar to the results on addition problems (see Appendix \ref{sec:appendix_subtract}). We do not include multiplication and division problems either, as LLMs perform poorly on these problems (even 5-digit multiplication yields an accuracy of about 0\%).}.
Let $a$ and $b$ be two randomly generated numbers, each question is formulated as:
\begin{verbatim}
Question: What is the sum of {a} and {b}? 
Answer: {a + b}
\end{verbatim}

\subsection{Probing Method}
\label{sec:probing}
\paragraph{Obtaining Hidden States.}
We choose the LLaMA-2 model family~\cite{touvron2023llama2} and Mistral-7B~\cite{jiang2023mistral} as base models to be investigated.
We feed the question text in Section \ref{sec:dataset} into the models, and save the hidden states of all layers.
For each layer, we obtain a set of hidden states (i.e. the residual stream) $\mathbf{H} \in \mathbb{R}^{n \times d_{model}}$ at every token position, where $n$ is the number of samples in the dataset. 

\paragraph{Training Probes.}
Following previous work, we adopt the widely acknowledged linear probing technique to reconstruct numbers from the hidden states.
To be specific, for each layer, given a set of hidden states $\mathbf{H}$ and their corresponding original numbers $\mathbf{X} = \{x\}$, we train a linear regressor $\mathcal{P}$ that yields best predictions $\mathbf{P} = \mathbf{H}\mathbf{W} + b$, where $\mathbf{W} \in \mathbb{R}^{d_{model}}$ and $b$ are the weights of $\mathcal{P}$.

In practice, directly performing linear regression could give erroneous results, as the value of numbers varies over a wide range.
We do a logarithmic operation on input numbers $\mathbf{X}$ with a base of 2 to guarantee the numerical stability of probes.

We utilize Ridge regression, which adds L2 regularization to the vanilla linear regression model, to construct the probes:
\begin{equation}
\small
    \mathbf{W}^*, b^* = \mathop{\arg\min}\limits_{\mathbf{W}, b} ||\log_{2}(\mathbf{X})-\mathbf{H}\mathbf{W}-b||^2_2 + \lambda ||\mathbf{W}||^2_2
\end{equation}
where $\mathbf{W}^*, b^*$ are the weights of regressors, and $\lambda$ is a hyperparameter that controls regularization strength.
In this way, we can predict logarithmic results $\mathbf{P}^* = \mathbf{H}\mathbf{W}^* + b^*$ based on the hidden states.

\begin{figure*}[ht]
    \centering
    \subfloat[$\rho$ of probes on $a$.]{
    \includegraphics[width=0.3\textwidth]{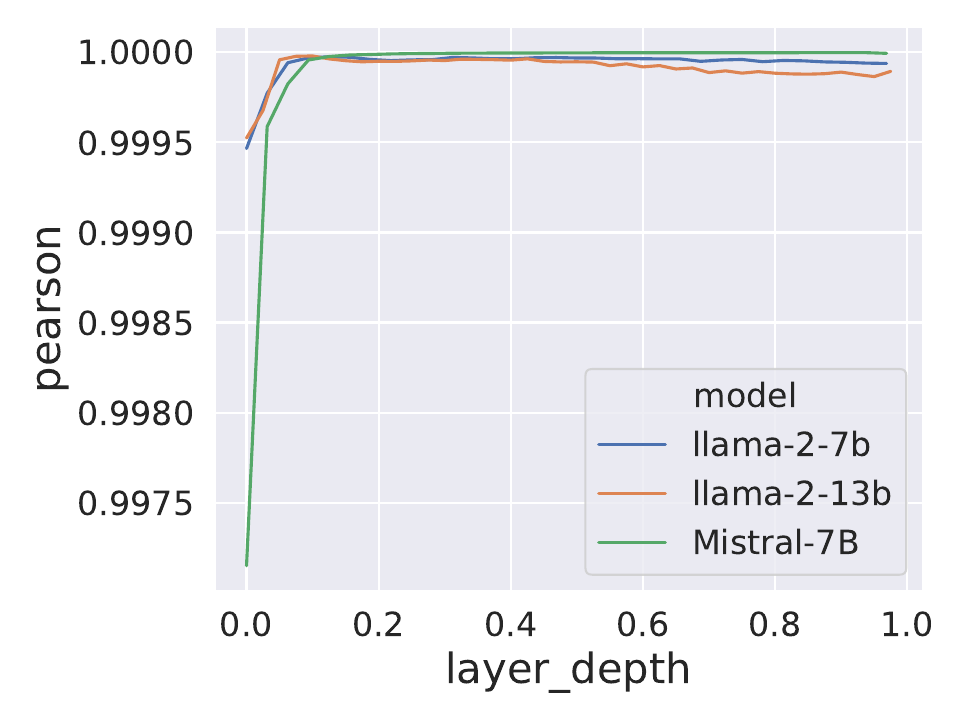}
    }
    \subfloat[$\rho$ of probes on $b$.]{
    \includegraphics[width=0.3\textwidth]{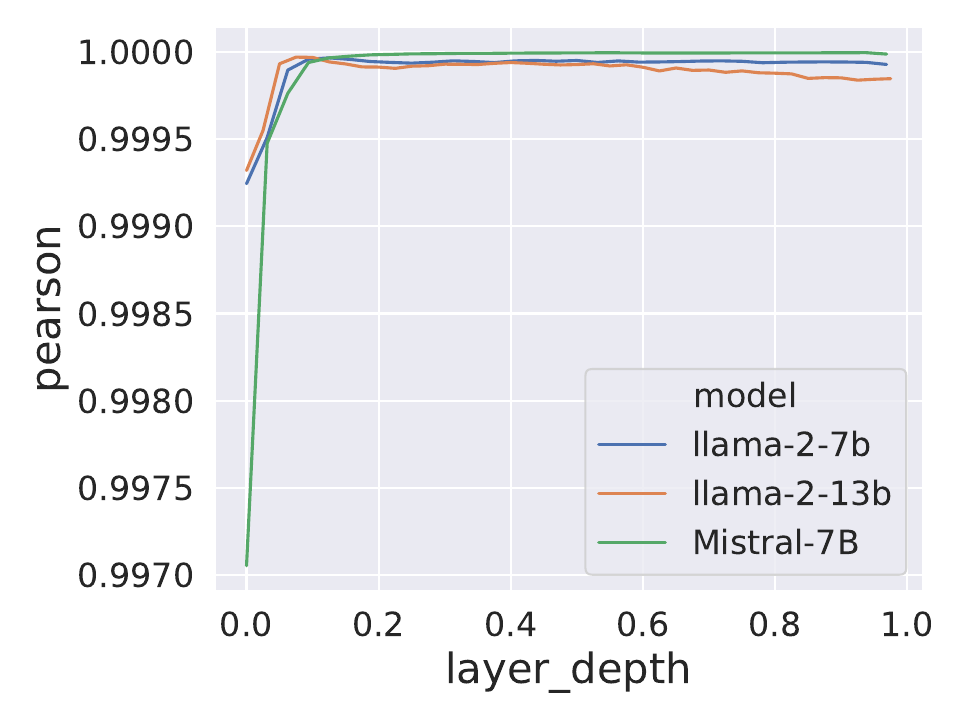}
    }
    \subfloat[$\rho$ of probes on $o$.]{
    \includegraphics[width=0.3\textwidth]{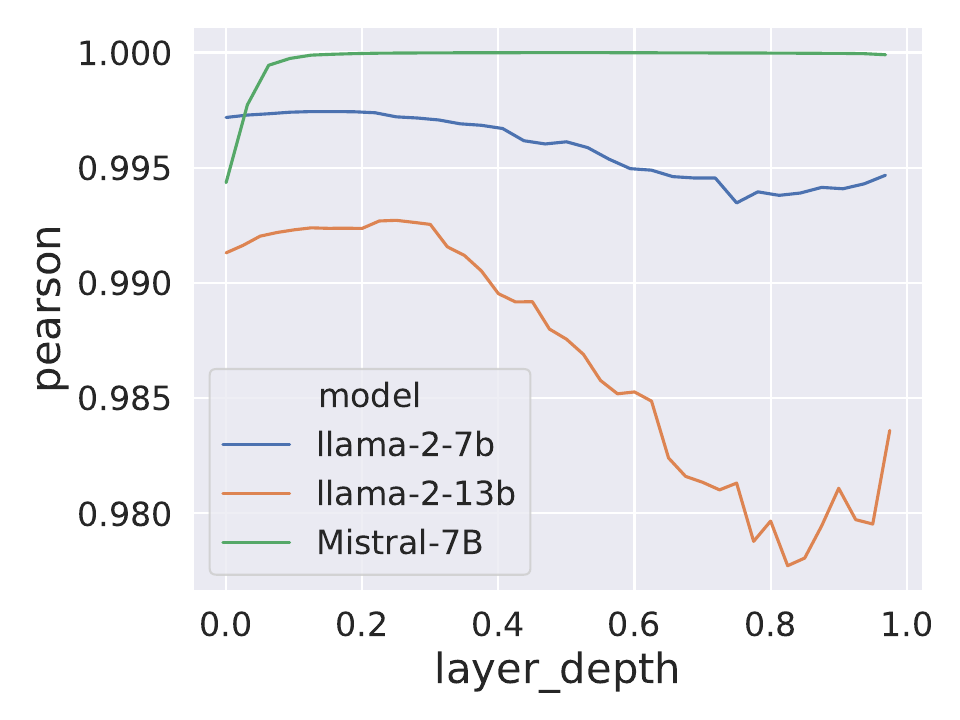}
    } \\
    \subfloat[$R^2$ of probes on $a$.]{
    \includegraphics[width=0.3\textwidth]{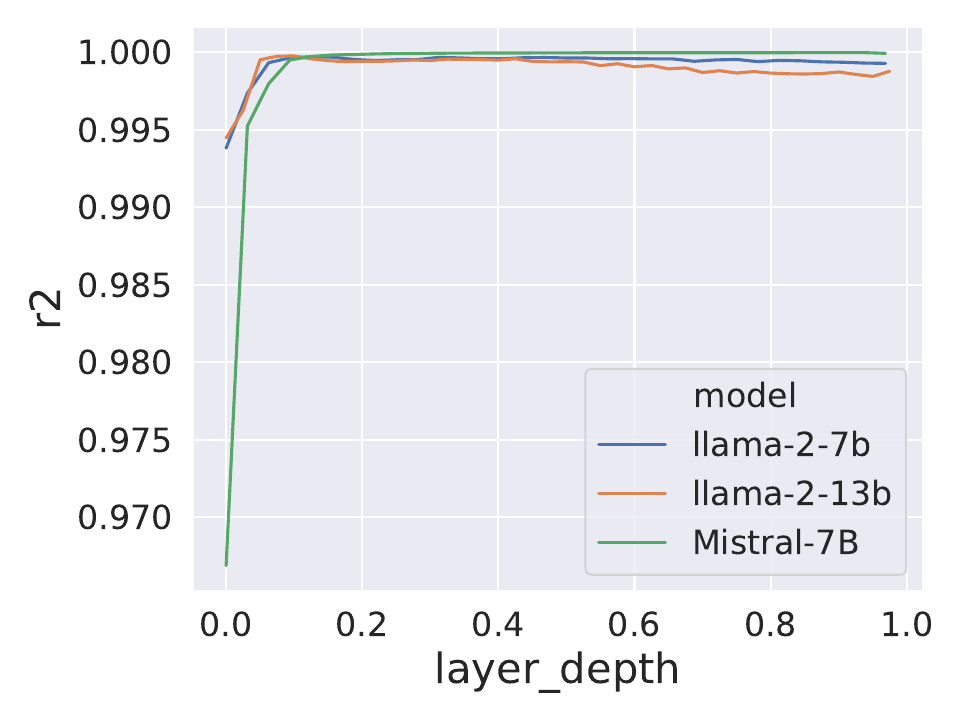}
    }
    \subfloat[$R^2$ of probes on $b$.]{
    \includegraphics[width=0.3\textwidth]{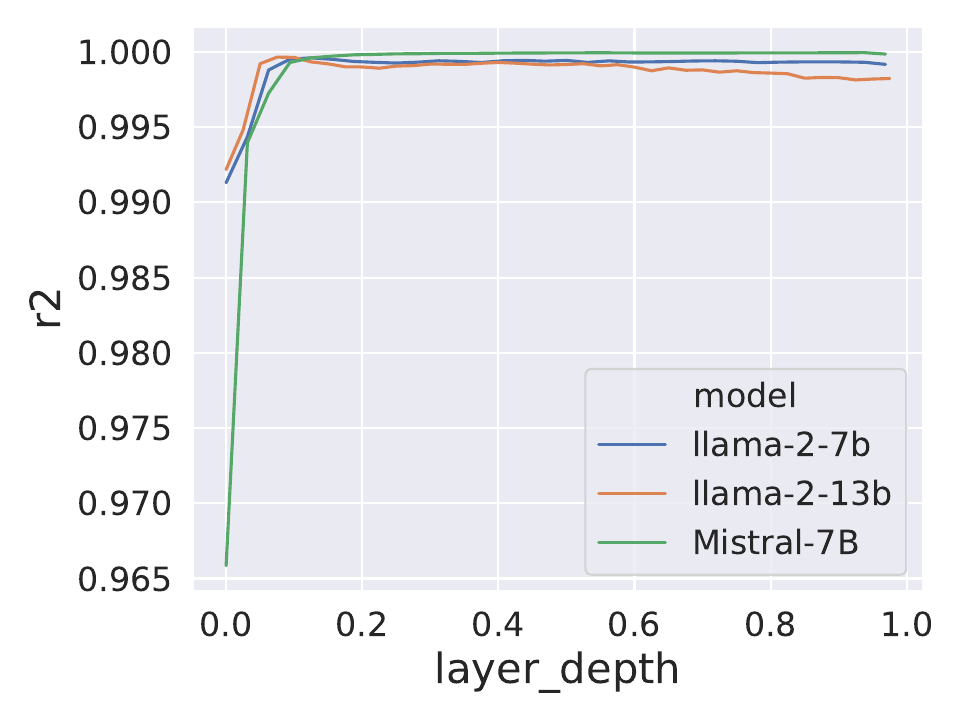}
    }
    \subfloat[$R^2$ of probes on $o$.]{
    \includegraphics[width=0.3\textwidth]{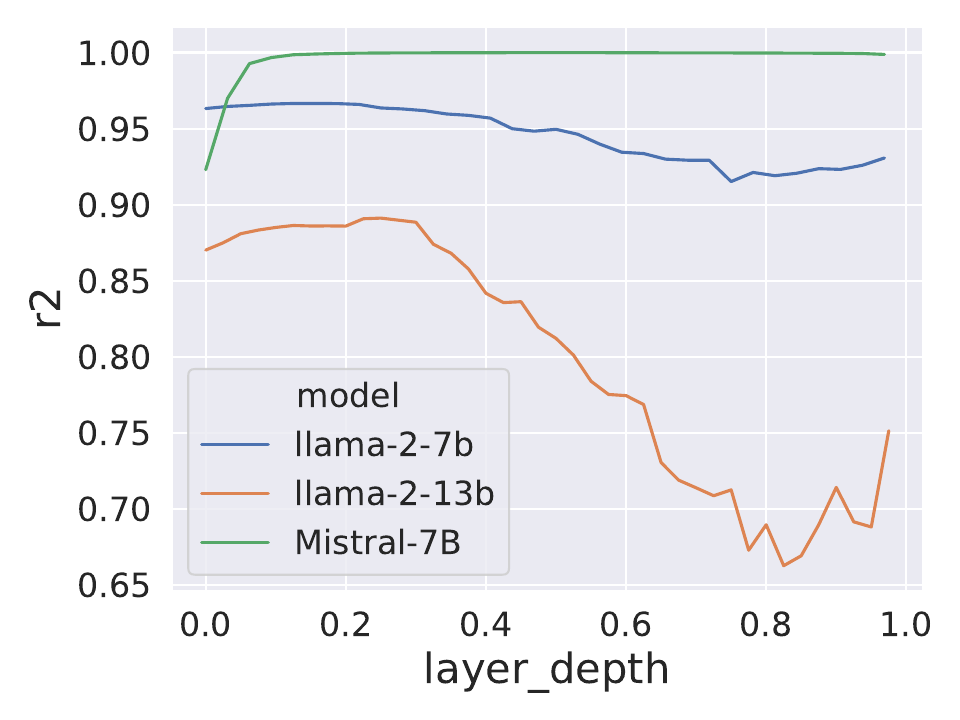}
    }
    \caption{Pearson coefficient ($\rho$) and out-of-sample $R^2$ of probes on different layers. $a$ and $b$ refer to the two input numbers denoted in Section \ref{sec:dataset}, and $o$ refers to the prediction of language models respectively. High $\rho$ and $R^2$ indicate the existence of encoded number values in the hidden states.}
    \label{fig:overall_correlation}
\end{figure*}

\subsection{Evaluation Metrics}
\label{ssec:metric}
We use two standard regression metrics on the probing task to evaluate the probes: $R^2$ which determines the proportion of variance in the dependent variable that can be explained by the independent variable, and the Pearson coefficient $\rho$ which measures the linear correlation between two variables.

As mathematical problems require a precise understanding of numbers, we introduce two additional metrics to examine how well can a language model encode numbers:

\textbf{Approximate accuracy (AAcc)} evaluates whether the predicted number is approximately the same as the original number, namely with an error margin of $<1\%$. 
Higher AAcc indicates that the number encoding is more likely to be precise.

\textbf{Mean square error (MSE)} is the average squared difference between probe predictions and actual values. 
Smaller MSE means lower loss during the encoding process.
\begin{align}
    \text{AAcc}(\mathbf{P}^*, \mathbf{X}) &= \frac{|(2^{\mathbf{P}^*} - \mathbf{X}) < 0.01X|}{|\mathbf{X}|} \\
    \text{MSE}(\mathbf{P}^*, \mathbf{X}) &= \text{avg}((\mathbf{P}^*-\log_{2}\mathbf{X})^2)
\end{align}

\subsection{Experimental Setup}
We use the original LLaMA-2-7B, LLaMA-2-13B, and Mistral-7B models without fine-tuning for all experiments.
The outputs are obtained by performing greedy search with a max new token restriction of 30 during decoding.
The regularization strength is set to $\lambda = 0.1$ for all probes.\footnote{See Appendix \ref{sec:appendix_implementation} for more details.}.

In main experiments, we probe 3 distinct values at different positions: the first number $a$ at the last digit of $a$ (for example, 3 for 123), the second number $b$ at the last digit of $b$, and the prediction of language models $o$ at the last token of the entire input text. 
We report the accuracy of $o$, i.e. the ratio of $o = a+b$, in Appendix \ref{sec:appendix_overall_accuracy}.

\begin{figure*}[ht]
    \centering
    \subfloat[AAcc of probes on $a$.]{
    \includegraphics[width=0.3\textwidth]{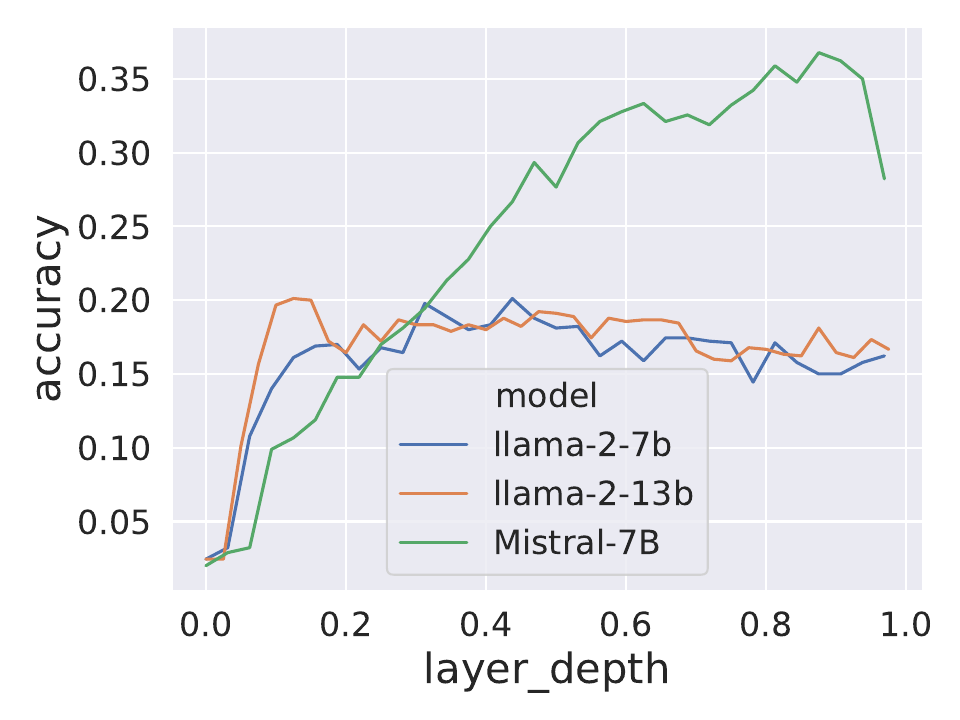}
    }
    \subfloat[AAcc of probes on $b$.]{
    \includegraphics[width=0.3\textwidth]{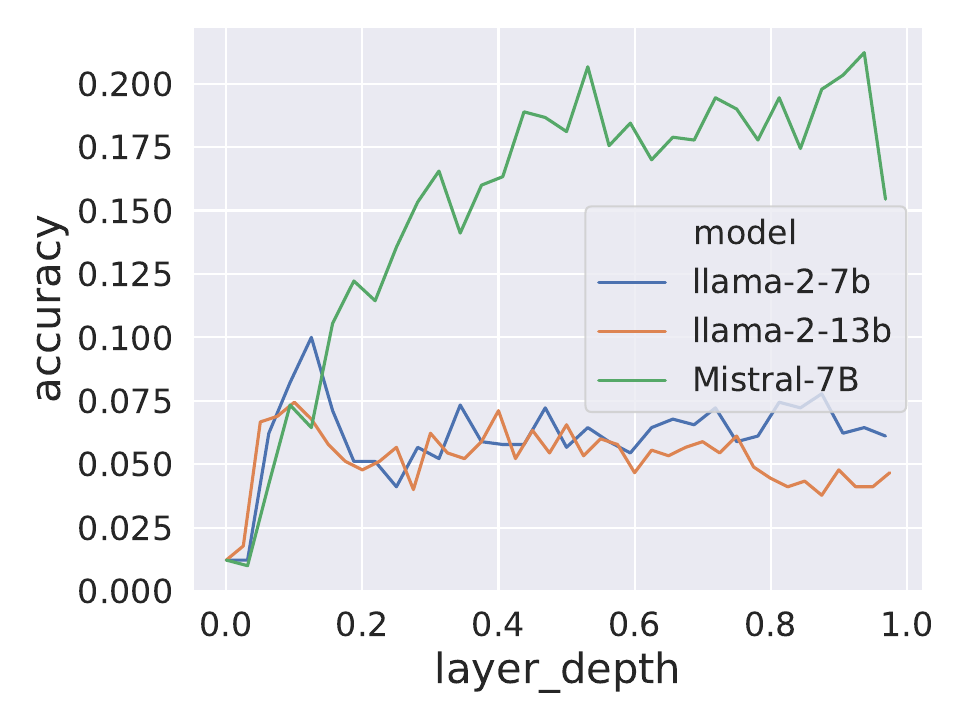}
    }
    \subfloat[AAcc of probes on $o$.]{
    \includegraphics[width=0.3\textwidth]{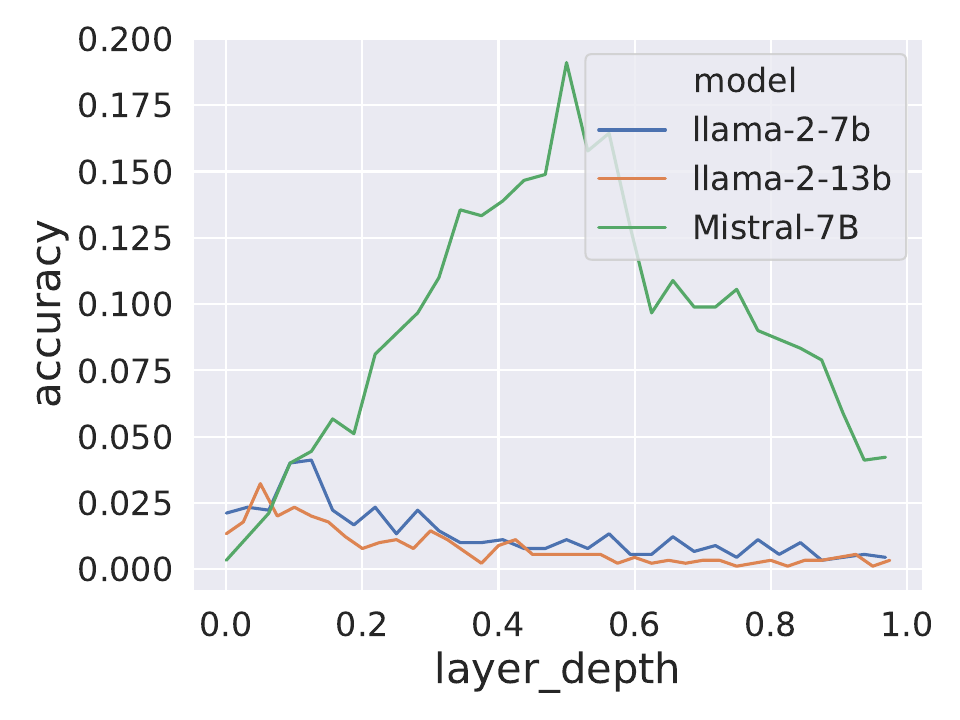}
    } \\
    \subfloat[MSE of probes on $a$.]{
    \includegraphics[width=0.3\textwidth]{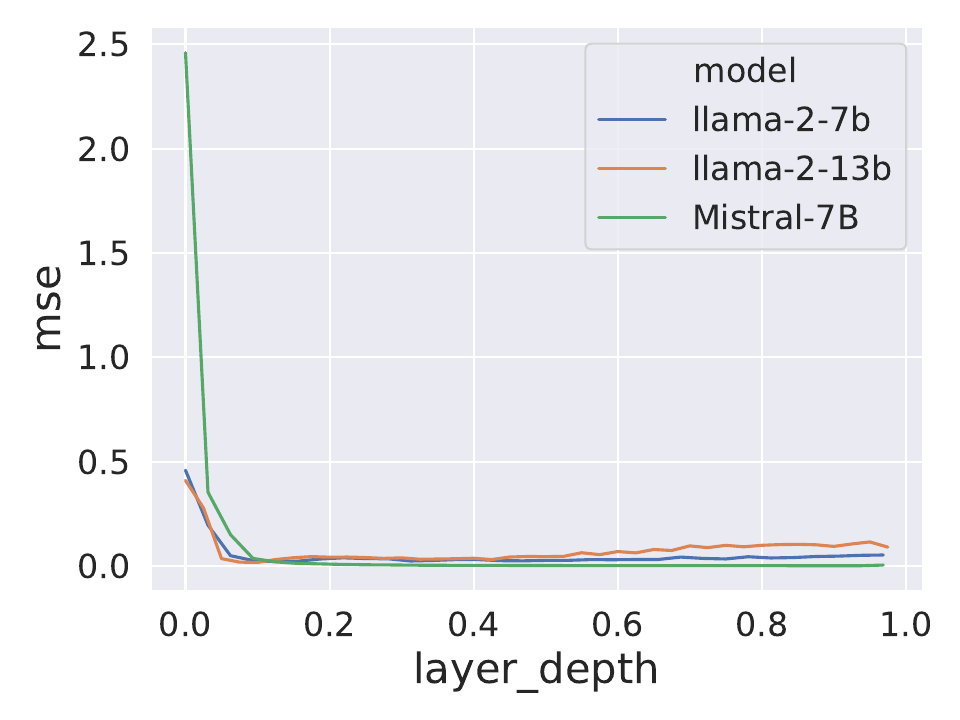}
    }
    \subfloat[MSE of probes on $b$.]{
    \includegraphics[width=0.3\textwidth]{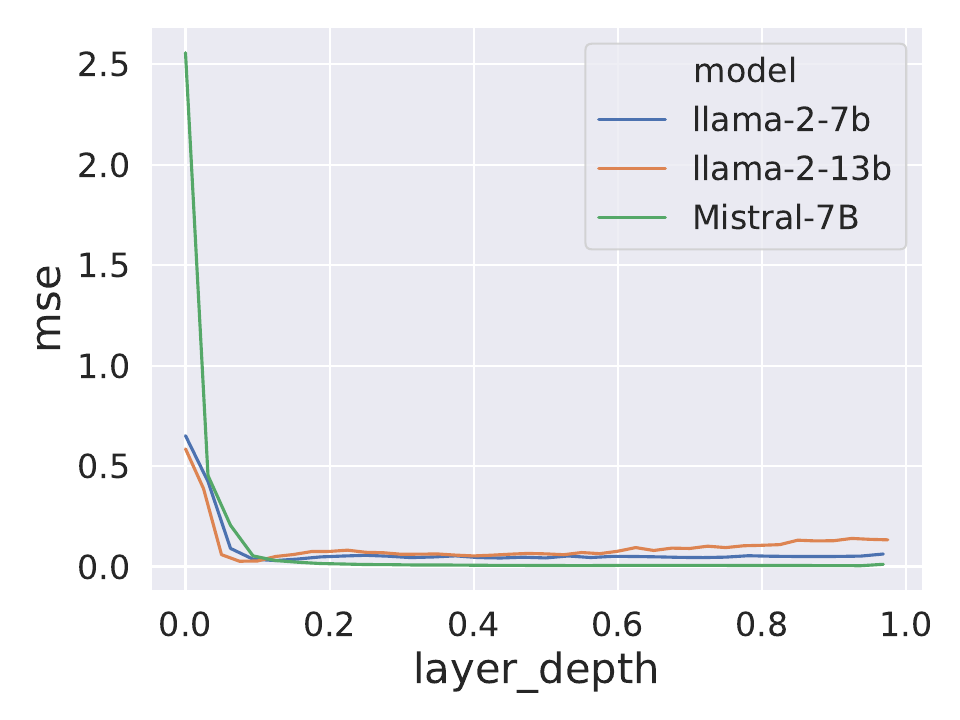}
    }
    \subfloat[MSE of probes on $o$.]{
    \includegraphics[width=0.3\textwidth]{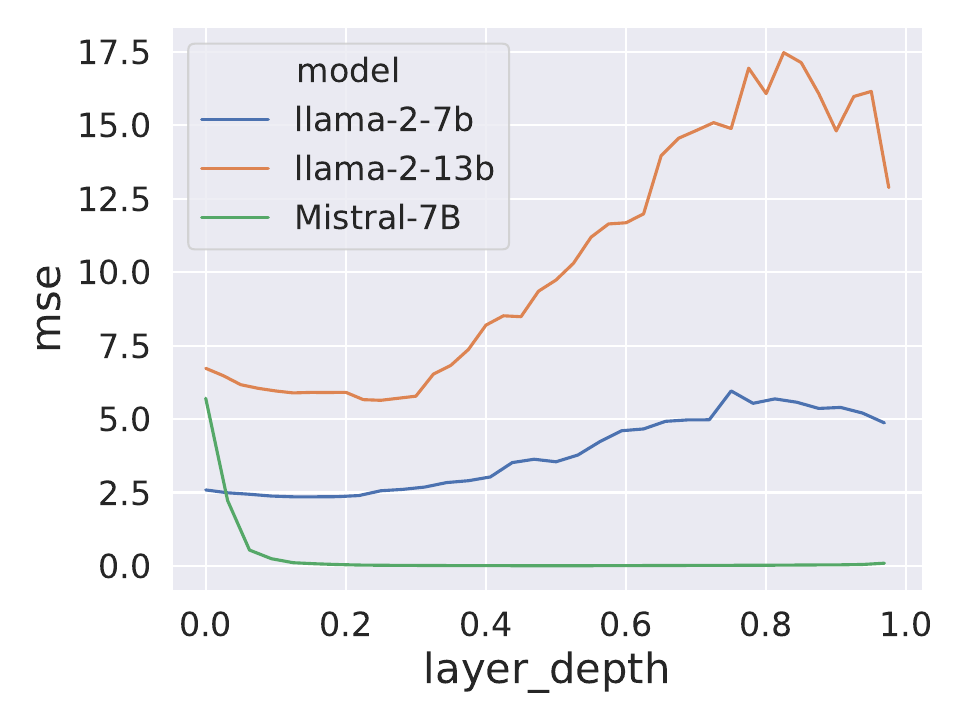}
    }
    \caption{Approximate accuracy (AAcc) and mean square error (MSE) of probes on different layers. $a$ and $b$ refer to the two input numbers denoted in Section \ref{sec:dataset}, and $o$ refers to the prediction of language models respectively. High AAcc and low MSE indicate precise number encoding.}
    \label{fig:overall_error}
\end{figure*}

\section{Do LLMs Encode Number Values?}
\subsection{The Existence of Encoded Number Values}
\label{ssec:number_existence}
\paragraph{LLMs do encode number values.}
We first inspect the overall Pearson coefficient ($\rho$) and out-of-sample $R^2$ on all layers.
High $\rho$ and $R^2$ indicate that LLMs are likely to be able to encode number values in their hidden states.
As illustrated in Figure \ref{fig:overall_correlation}, the probes achieve surprisingly high $\rho$ and $R^2$ on all layers, proving that the hidden states of LLMs contain the encoded value of input numbers, and the encoding process starts from even the first layer.
Meanwhile, notice that both $\rho$ and $R^2$ slightly drop on late layers, which may indicate that intermediate layers better encode number values.

\paragraph{Linear probes cannot reconstruct the precise value.}
Aside from the existence of encoded number values, we are also interested in their precision, which is depicted by AAcc and MSE in Figure \ref{fig:overall_error}.

In contrast to high correlation coefficients, the AAcc is below 50\% on all layers, which means that the linear probes have difficulty in precisely reconstructing the input numbers.
The trends in AAcc and MSE are consistent with $\rho$ and $R^2$, indicating that LLaMA-2 models achieve the most precise number encoding in intermediate layers, but the encoding faces more error in deeper layers.

This phenomenon may indicate that language models use stronger non-linear encoding systems, which we will further explore in Section \ref{ssec:linearity};
Or it may be a hint that the number encoding in language models is not precise\footnote{See Appendix \ref{sec:appendix_partial} for more detailed experiments.}.

\begin{figure*}[ht]
    \centering
    \subfloat[Layer 0]{
    \includegraphics[width=0.3\textwidth]{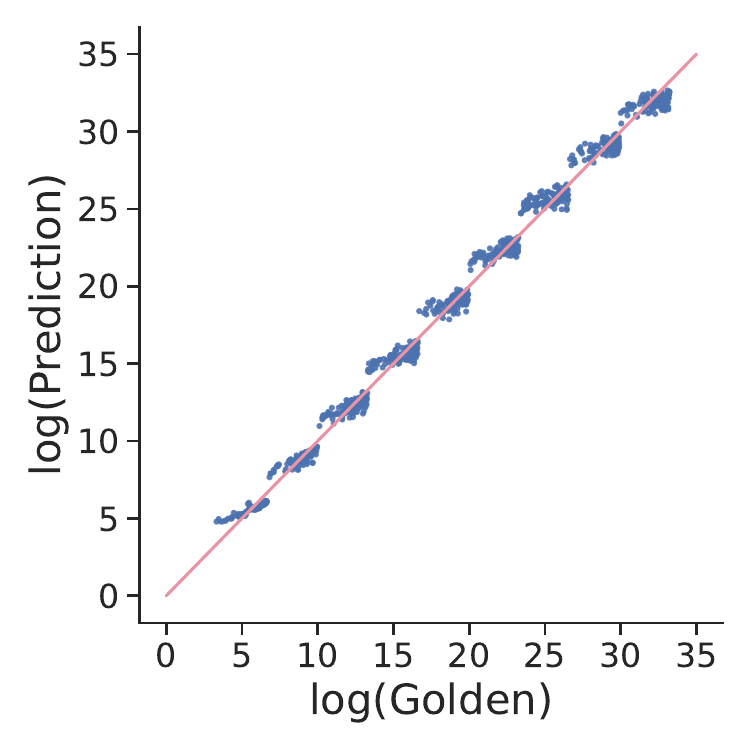}
    }
    \subfloat[Layer 10]{
    \includegraphics[width=0.3\textwidth]{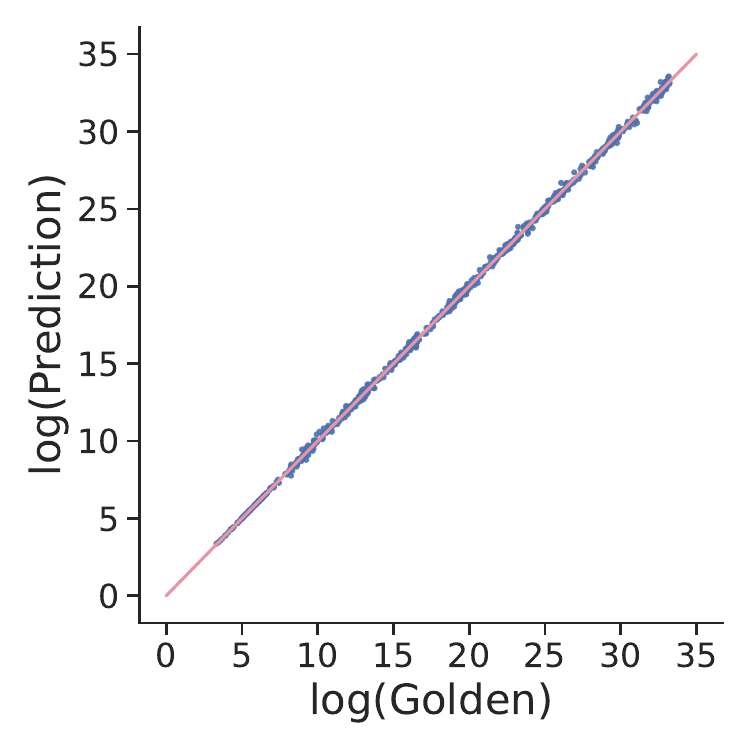}
    }
    \subfloat[Layer 30]{
    \includegraphics[width=0.3\textwidth]{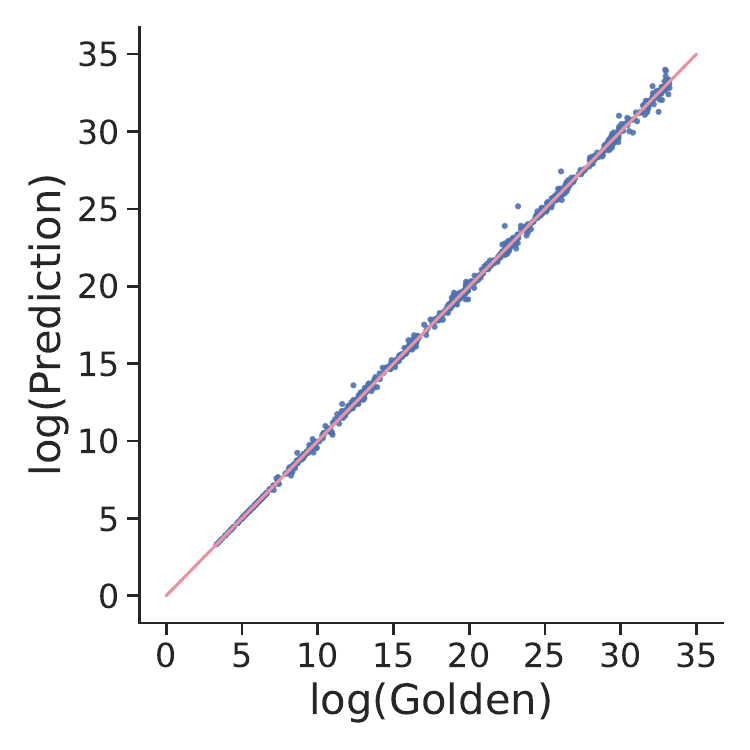}
    }
    \caption{How the pattern of probe predictions on the first input number $a$ changes as the layer gets deeper. Probe predictions on different layers of LLaMA-2-7B show different patterns.}
    \label{fig:evolution}
\end{figure*}

\begin{figure*}[ht]
    \centering
    \subfloat[MSE of probes on $a$.]{
    \includegraphics[width=0.4\textwidth]{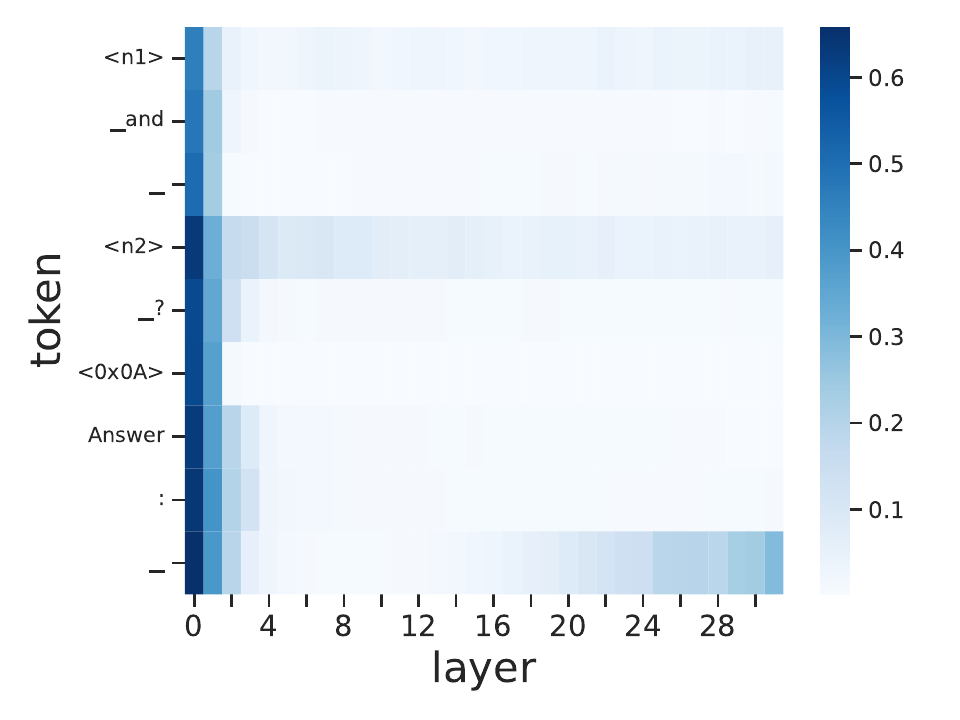}
    }
    \subfloat[MSE of of probes on $b$.]{
    \includegraphics[width=0.4\textwidth]{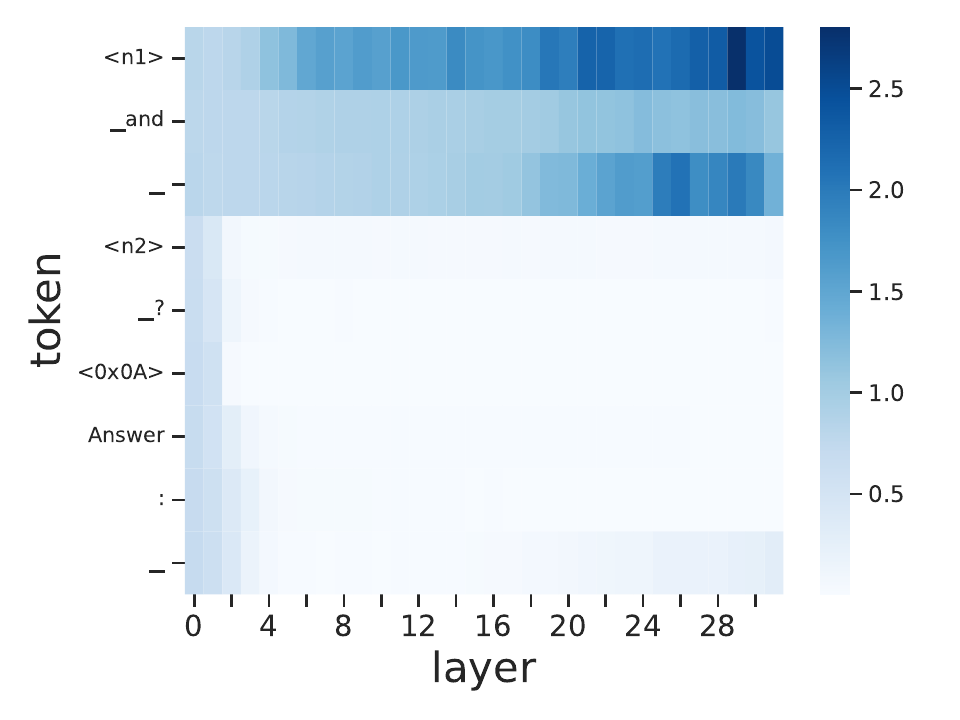}
    }
    \caption{The mean square error (MSE) of probes at different token positions on LLaMA-2-7B. \texttt{<n1>} represents the last token of the first input number $a$, and \texttt{<n2>} represents the last token of the second input number $b$, respectively. The rectangular pattern indicates that the value of an input number can be read out at any subsequent position.}
    \label{fig:persistence}
\end{figure*}

\subsection{Number Encoding Patterns are Different across Layers}
To better analyze how language models encode numbers, we pick distinct layers in LLaMA-2-7B and observe how the pattern of probe predictions changes as the layer gets deeper.
Layer 0 (i.e. the first transformer block after embedding layer), 10, and 30 are selected to represent early, intermediate, and late layers respectively.
The trend of change on the first input number $a$ is shown in Figure \ref{fig:evolution}.

On early layers like layer 0, the predictions of probes are distorted to some extent:
for original numbers with the same length, their corresponding predictions in the figure display a pattern of horizontal lines.
This phenomenon indicates that early layers focus on the length of numbers, which corresponds to the number of input digit tokens.

As the layer gets deeper, probes on intermediate layers show the best performance.
On layer 10, the predicted results are very close to the actual answers, yielding a near-perfect linear probe for original numbers.
However, noise emerges in the prediction results again in late layers, with the form of uniformly distributed errors.

The trend of change leads us to a conjecture that language models first roughly estimate the value of a number with its token length, and then refine the estimation in subsequent layers.
The process may not be lossless, which leads to errors in the final number encoding of language models.

\subsection{Numeric Information Persist at Subsequent Positions}
\label{ssec:persistence}
Another question is whether these encoded values are only stored at certain positions, or are they persist at subsequent positions.
For input number values $a$, $b$, we train probes at every individual token position to examine where these values exist.
Figure \ref{fig:persistence} shows the mean square error of probes on the LLaMA-2-7B model.

The results demonstrate a clear rectangular pattern, indicating that the value of an input number can be read out at any subsequent position.
In other words, the number values would persist at subsequent positions.
It is also worth noticing that the probing accuracy on the last token is lower than other positions, which may be interpreted as language models do not continue to remember input numbers after computing the final outcome.

\begin{figure}[ht]
    \centering
    \includegraphics[width=0.4\textwidth]{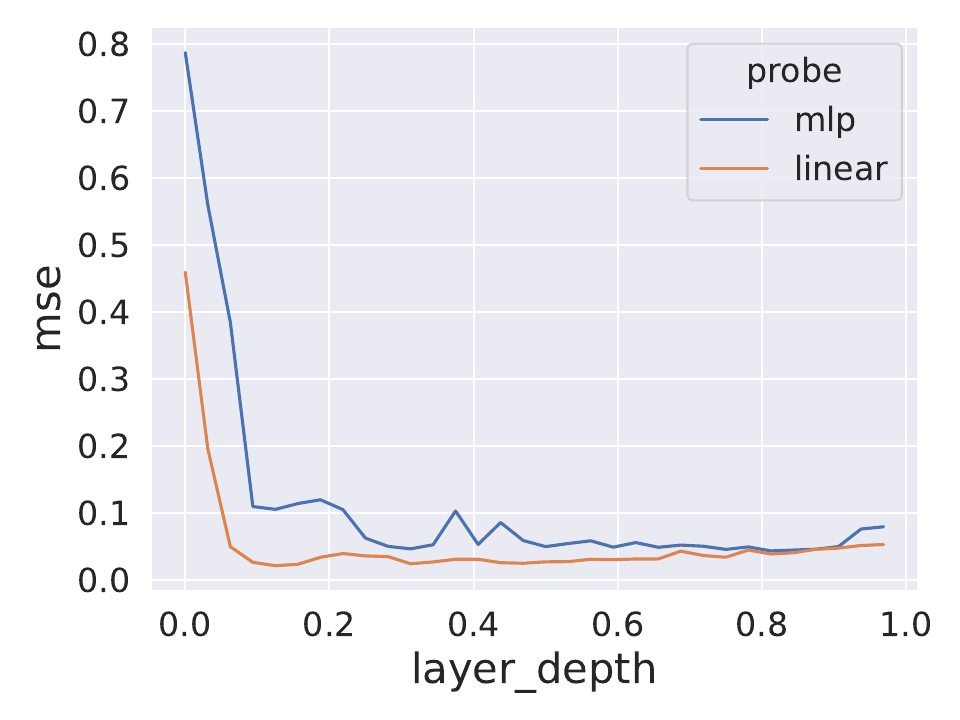}
    \caption{The comparison between linear probes and MLP probes on mean square error (MSE). The MLP probes do not show advantage over linear probes. More detailed experiments are reported in Appendix \ref{sec:appendix_linearity}.}
    \label{fig:linear}
\end{figure}

\subsection{LLMs Encode Numbers Linearly}
\label{ssec:linearity}
Previous work~\cite{nanda2023emergent, gurnee2023language} on probing neural networks propose the linear representation hypothesis: the presence of features of a neural network can be proved by training a linear projector which projects the activation vector to the feature space, and complex structures are unnecessary.
To verify whether the numbers can be represented linearly, we follow the method of \citet{gurnee2023language} which trains two-layer MLP probes and compares their performance with linear probes.
The MLP probes have an intermediate hidden state of 256 dimensions and can be formulated as:
\begin{equation}
    \mathbf{P} = \mathbf{W}_{2}\text{ReLU}(\mathbf{W}_{1}\mathbf{H}+b_{1})+b_{2}
\end{equation}
where $\mathbf{W}_{1}, \mathbf{W}_{2}, b_{1}$ and $b_{2}$ are trainable weights.

Figure \ref{fig:linear} demonstrates the comparison between MLP probes and linear probes on mean square error.
We find that nonlinear MLP probes do not show any clear advantage over linear probes, proving that the encoded number values can be represented linearly, or at least near-linearly.

\section{Do LLMs Utilize Number Values?}
The previous section has proved the existence of encoded number values in language models.
However, an inherent issue is that the probed information is only correlational to the output of models, and no causal effects can be directly claimed~\cite{belinkov2022probing}.

In this section, we will try to verify the hypothesis that language models do use the encoded number values to get their calculation results by performing a set of intervention experiments. 
Given an input question $Q$ with an expected result of $o$, we intervene in the internal activation of language models to make it believe in an altered question $Q'$, and observe how the new result $o'$ changes.

To ensure the effectiveness of the intervention, we conduct the experiments on Mistral-7B with 4-digit addition questions as input, where the model could correctly answer most of the questions.
\begin{figure}[htp]
    \centering
    \includegraphics[width=0.4\textwidth]{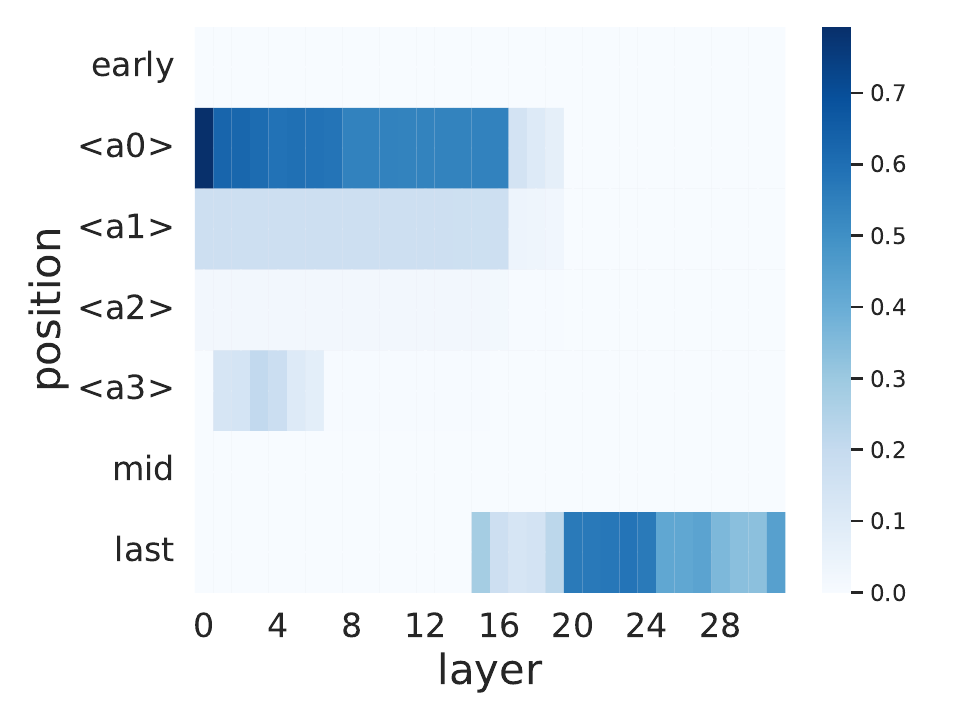}
    \caption{The effect of patching on different components. Early and mid refer to the non-number tokens before and after the first input number $a$, and last refers to the last token of the input text.}
    \label{fig:intervene_patch}
\end{figure}

\subsection{Patching Encoded Numbers}
\label{ssec:activation_patching}
Firstly, we study the influence of number encoding at different positions by changing the activation of language models.
We adopt the activation patching technique proposed by \citet{stolfo2023mechanistic} to quantify the importance of encoded number values $\mathbf{h}_i$ at different layers $i$ and different token positions.

To be specific, given an input addition problem consisting of input numbers $a$ and $b$, we will conduct the following procedure:
\begin{enumerate}
    \item Obtain the clean output of the language model $o = LM(a, b)$.
    \item Replace $a$ with another number $a'$ to get a new output $o' = LM(a', b)$, and record the hidden states $\mathbf{h}'$ at certain position $t$ during the forward pass;
    \item Perform an additional forward pass with $a$ and $b$ as input numbers, where we substitute the hidden state $\mathbf{h}_i$ of layer $i$ with $\mathbf{h}'_i$. This would lead to an intervened result $o^{*}$. 
\end{enumerate}
We set $a' = 9999$ in our experiments, and evaluate the effect of intervention as:
\begin{equation}
    E(i, t) = \frac{|o^{*}-o|}{|o'-o|}
\end{equation}
which measures how much a specific layer $i$ at position $t$ affects the final result.
Note that the metric is intended for qualitative rather than quantitative analysis.

Figure \ref{fig:intervene_patch} demonstrates the effect of activation patching on different components, from which we can draw multiple observations:

\begin{table}[htbp]
    \centering
    \begin{tabular}{c|c|p{0.4\linewidth}}
    \hline
    Patching & Result & Explanation \\
    \hline
    None & 6912 & 5678+1234=6912 \\
    Full & 11233 & 9999+1234=11233 \\
    $5 \xrightarrow{} 9$ & 10912 & 9678+1234=10912 \\
    $6 \xrightarrow{} 9$ & 7212 & 5978+1234=7212 \\
    $7 \xrightarrow{} 9$ & 6932 & 5698+1234=6932 \\
    $8 \xrightarrow{} 9$ & 6913 & 5679+1234=6913 \\
    \hline
    \end{tabular}
    \caption{Patching results on the question ``Question: What is the sum of 5678 and 1234 ?'' by patching the activation on layer 8.}
    \label{tab:patch_example}
\end{table}

\paragraph{Each digit affects the result independently.}
The effect of patching on different number digits displays a clear pattern: the earlier a digit appears, the more patching it changes the final output value.
While the latter digits encode the values of partial number sequences (See Appendix \ref{sec:appendix_partial} for details), activation patching seems to only change the final result by the value of the digit itself.
For example, although the activation at digit ``3'' in ``1234'' encodes the value of 123, patching it equals changing the input number to "1294" rather than "9994", as demonstrated in Table \ref{tab:patch_example}.
More detailed experiments are reported in Appendix \ref{sec:appendix_patch}. 

\paragraph{Language models concern only certain tokens during calculation.}
Despite our finding in Section \ref{ssec:persistence} that encoded number values would persist in subsequent tokens, patching non-number tokens has almost zero effect on the final outcome. 
This pattern indicates that the encoded number values at most positions are simply ``memorized'' rather than  ``used'' by the language model.
An exception is the last token, where language models seem to store their calculation results.

\paragraph{Early and late layers play different roles.}
The effect of activation patching can be divided into two parts: 
on early layers before layer 14, patching the number tokens greatly influences the final outcome, while patching the last token is mostly ineffective; but in late layers after layer 20 it is just the opposite.
We assume that early layers perform the task of processing the value of input number token sequences, while late layers use encoded values to calculate the final outcome, which is similar to the findings in \citet{stolfo2023mechanistic}.

\begin{figure}[t]
    \centering
    \includegraphics[width=0.42\textwidth]{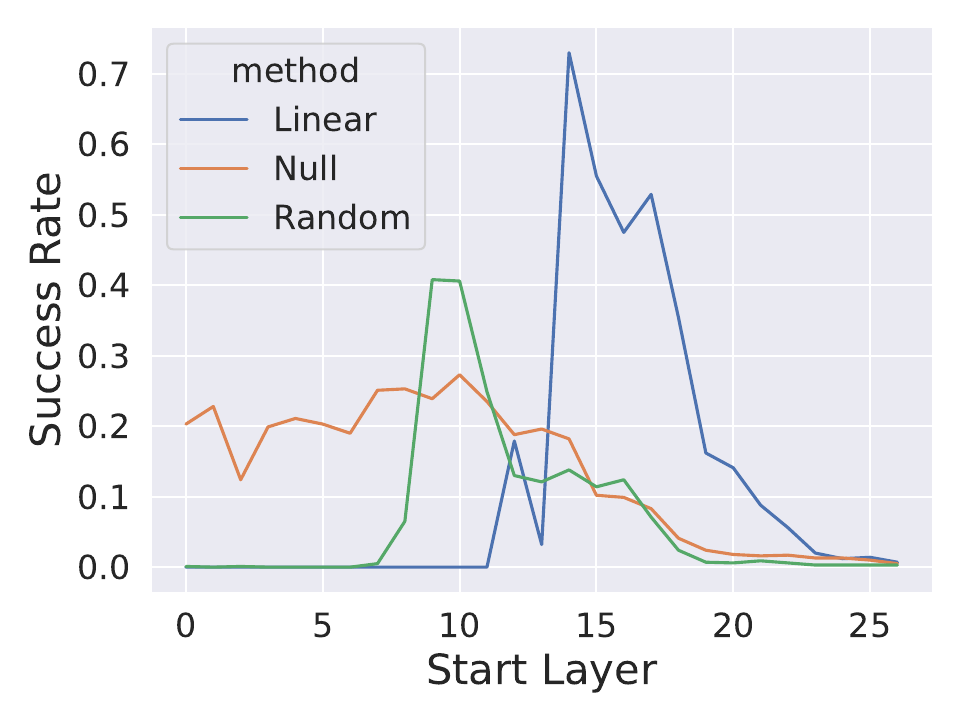}
    \caption{The success rate of performing a linear intervention on 6 consecutive layers. More detailed experiments are reported in Appendix \ref{sec:appendix_linear}.}
    \label{fig:intervene_linear}
\end{figure}

\subsection{Linearly Intervening Encoded Numbers}
To determine whether the encoded computational results causally affect the outcome of language models, we linearly intervene the hidden states and see whether the output changes as expected.

\paragraph{Method.}
Following the method of \citet{nanda2023emergent}, for each intervened layer $i$, we add the number encoding direction vector $\mathbf{d}_i$ to the residual stream $\mathbf{h}_i$:
\begin{equation}
    \mathbf{h}_i^{'} = \mathbf{h}_i + \alpha \mathbf{d}_i
\end{equation}
where $\alpha > 0$ is a scaling factor and the direction vector $\mathbf{d}_i$ is obtained by normalizing the probe coefficients:
\begin{equation}
    \mathbf{d}_i = \frac{\mathbf{W}_i}{|\mathbf{W}_i|}
\end{equation}
Considering that the probed number value is the projection of $\mathbf{h}_i$ along the direction $\mathbf{d}_i$, the effect of our intervention is to ``push'' the residual stream towards a larger encoded number.
We set $\alpha = 2$ in our experiments, and intervened language models outputting a larger number $o' > o$ than the original prediction $o$ is viewed as a success.

In the linear intervention experiment, we choose probes for language model predictions $o$ at the last input token to obtain the direction vector $\mathbf{d}_i$, and perform an intervention on every newly generated token.
We use a test set of 1,000 entries and measure the efficacy of our intervention by observing the ratio of successful interventions.

We also use two alternative directions as baselines: normalized $\mathbf{h}_i$ as null intervention, and a random unit vector $\mathbf{I}$ as random intervention.

\paragraph{Result and Findings.}
Figure \ref{fig:intervene_linear} shows the success rate of intervening on 6 consecutive layers.
Linear intervention achieves the highest success rate of 0.73 when intervening layer 14 to layer 19, outperforming the null intervention baseline by a large margin.
This suggests that the linearly encoded number values are causal to model predictions.

It is also worth noticing that intervening on mid-late layers is significantly more effective than on early layers and late layers.
We hypothesize that this phenomenon is related to the findings of~\citet{stolfo2023mechanistic}: language models use mid-late layers to perform arithmetic computations, while the late layers are responsible for converting the computational result to output tokens.

\section{Discussion and Future Directions}
In previous sections, we find that LLMs know the value of numbers and utilize the encoded number values to perform calculations.
However, the compression may not be lossless, and the calculation ability scales with model size.
Moreover, the ability to understand and utilize numbers is positively correlated to mathematical competency.
These findings reveal some future research directions that are potentially promising.

\paragraph{The exact way that LLMs encode numbers.}
While our experiments show that the original input number cannot be reconstructed from the hidden state via linear probes, there exists a possibility that the LLMs encode numbers in a way that is close to a linear projection but not identical, such as the floating-point system~\cite{muller2018handbook}.
Finding out the exact encoding, if possible, will give us a better insight into how LLMs function.

\paragraph{Specialized number encoding systems.}
The loss of encoded number values in LLMs will inevitably bring errors to subsequent computation, especially for large input numbers.
Developing specialized encoding systems that could give precise presentations for numbers~\cite{golkar2023xval} could eliminate errors at the root, thus helping LLMs better solve mathematical problems.

\paragraph{Mitigating computational errors with encoded numbers.}
By adding modules that directly utilize the encoded numbers in language models, the computational errors may be further reduced, especially on large-number calculations.
We conduct a pioneer experiment in Appendix \ref{sec:appendix_direct} to reveal the potential of controlling computational errors with probed numbers.

\section{Related Work}
\paragraph{Large Language Models on Mathematical Problems.}
Large language models (LLMs) like the GPT series~\cite{openai2023gpt4}, PaLM~\cite{anil2023palm} and LLaMA~\cite{touvron2023llama, touvron2023llama2} have demonstrated their impressive ability in various fields~\cite{zhao2023verify, li2023chain, taori2023stanford, chen2021evaluating, nijkamp2022codegen, li2023starcoder}.
On mathematical datasets like GSM8K~\cite{cobbe2021training} and MATH~\cite{hendrycks2021measuring}, there have been methods like chain-of-thought reasoning~\cite{wei2022chain} and self-consistency~\cite{wang2022self} to help LLMs better solve these questions.
Specialized large language models like MetaMath~\cite{yu2023metamath} and Math-Shepherd~\cite{wang2023math} also show great competency.

\paragraph{Interpreting Internal Representations in Language Models.}
Prior research has unveiled that language models are able to store certain information in their hidden states, for example, passive voice~\cite{shi2016does} and sentence structure~\cite{tenney2018you}.
By adopting the probing technique~\cite{alain2016understanding, belinkov2022probing}, complex representations have also been detected in language models:
\citet{li2022emergent} shows that language models are capable of memorizing the state of an Othello game, and \citet{nanda2023emergent} further proves that the states can be linearly represented;
\citet{li2021implicit} claims that language models are able to encode the properties and relations of entities;
\citet{gurnee2023language} reveals evidence that large language models build spatial and temporal representations about an entity from early layers.

\paragraph{Explaining Numbers and Arithmetic in Language Models.}
How language models process numbers has been studied by multiple researchers.
\citet{wallace2019nlp} detects the existence of numeracy in static pre-trained word embeddings.
\citet{hanna2024does} finds a critical circuit that performs greater-than comparisions in GPT-2.
\citet{stolfo2023mechanistic} studies how language models process arithmetic information by intervening on specific modules of the model.
\citet{zhong2024clock, engels2024not} discover evidence that numbers on modular arithmetic may be circularly encoded.

\section{Conclusion}
In this paper, we study the question of whether and how large language models encode the value of numbers.
If number values can be extracted from the internal representations of LLMs, we can assume that LLMs encode the value of numbers in their hidden states.
We construct a dataset consisting of simple addition problems and introduce linear probes to investigate whether language models encode number values.

Experimental results prove that LLMs do encode the value of input numbers, and the representation could be linearly read out.
The ability to linearly encode numbers is consistent across different model scales, and the encoding seems to be the most precise on intermediate layers.
Further experiments show that LLMs utilize the encoded number values to perform arithmetic calculations, and the behavior of language models can be controlled via simple linear interventions, proving the causal connection between encoded numbers and model outputs.

Our work shows a glimpse of the internal mechanisms of how language models solve mathematical questions.
Future works on the internal representations of numbers, for example, better probes and specialized number encoders, may enhance the mathematical competence of language models in an explainable way.

\section*{Limitations and Risks}
While we explore the inner mechanisms of how language models understand numbers, the probes trained in our current method are only approximations of the encoded numbers rather than exact internal presentations.
Directly performing calculations with probes would lead to undesired results.
Meanwhile, our experiments are conducted on LLMs whose parameters are openly available, while other LLMs the ChatGPT or GPT-4 may exhibit different behaviors.

\bibliography{custom}

\appendix

\section{Dataset Details}
\label{sec:appendix_dataset}
The dataset in Section \ref{sec:dataset} contains 9000 addition problems.
For each number of digits between 2 and 10, 1000 problems are generated, and two numbers in the same problem share the same digit.
For questions whose number has 4 or fewer digits, we list all possible combinations of numbers and randomly sample 1,000 of them to generate the questions.
For questions whose number has 5 or more digits, we randomly sample both numbers to generate the 1000 questions.

\section{Experiment Implementation}
\label{sec:appendix_implementation}
The experiments are conducted on 4 NVIDIA GTX 3090 GPUs.
Acquiring the hidden states of LLMs on our synthetic dataset requires 10\~20 GPU hours per model.

We obtain the LLaMA-2 models and Mistral-7B model from the huggingface model hub, and implement the experiments with the huggingface transformers Python library.
The probes are trained with the scikit-learn Python library.
We use the TransformerLens library\footnote{\url{https://github.com/neelnanda-io/TransformerLens}} for intervention experiments.
We follow the terms of use of all models and use them only for research.

\begin{figure}[ht]
    \centering
    \includegraphics[width=0.45\textwidth]{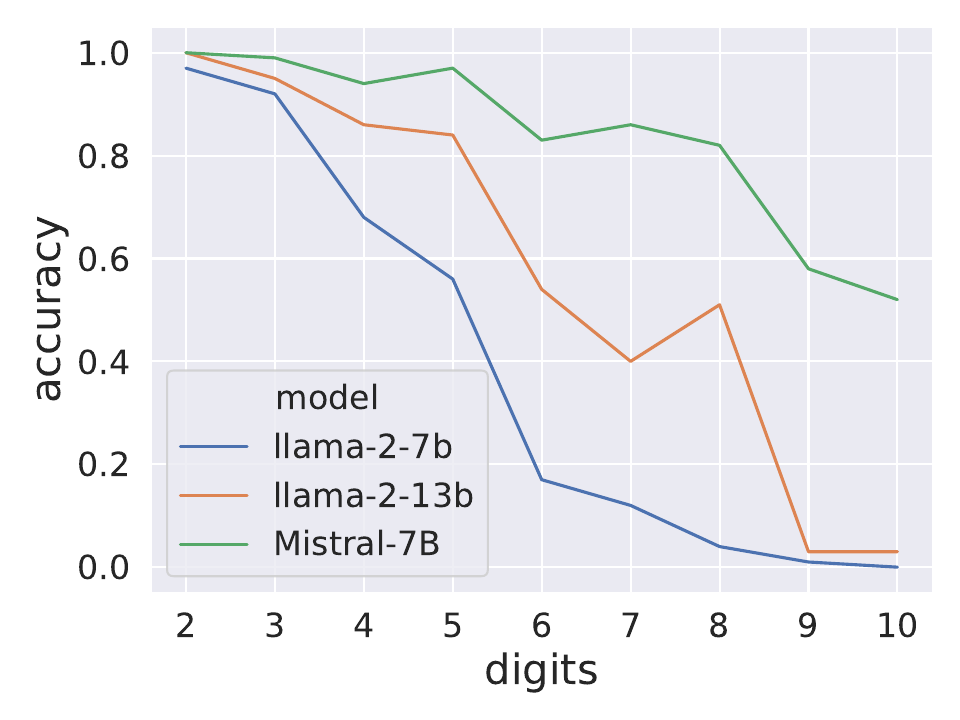}
    \caption{The overall accuracy of language model predictions on addition problems.}
    \label{fig:appendix_overall_accuracy}
\end{figure}

\begin{figure*}[htbp]
    \centering
    \subfloat[$\rho$ of probes on $a$.]{
    \includegraphics[width=0.3\textwidth]{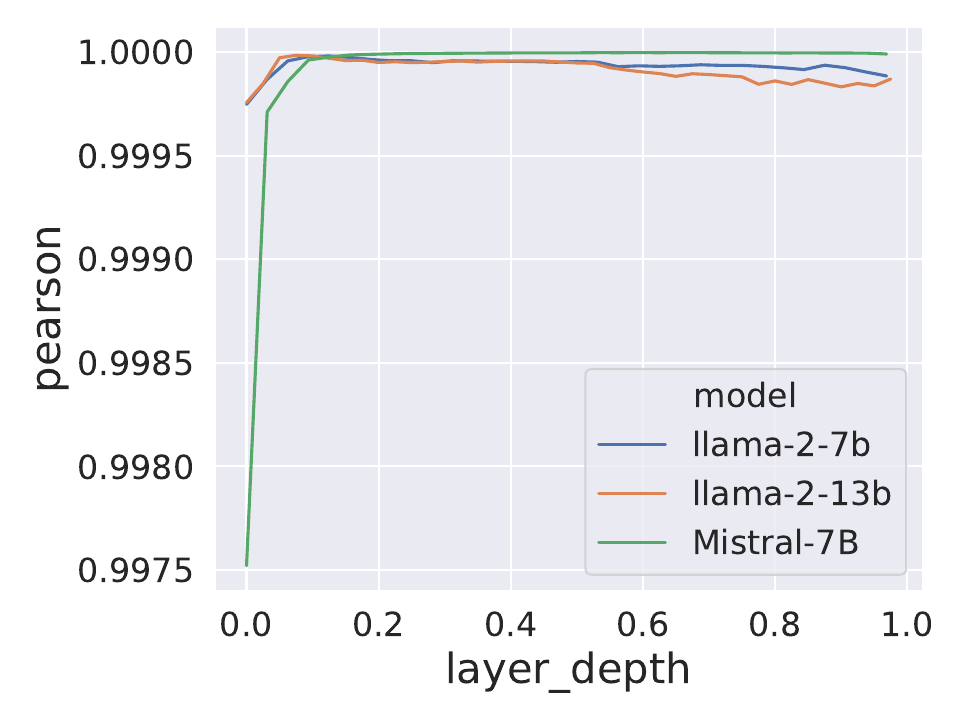}
    }
    \subfloat[$\rho$ of probes on $b$.]{
    \includegraphics[width=0.3\textwidth]{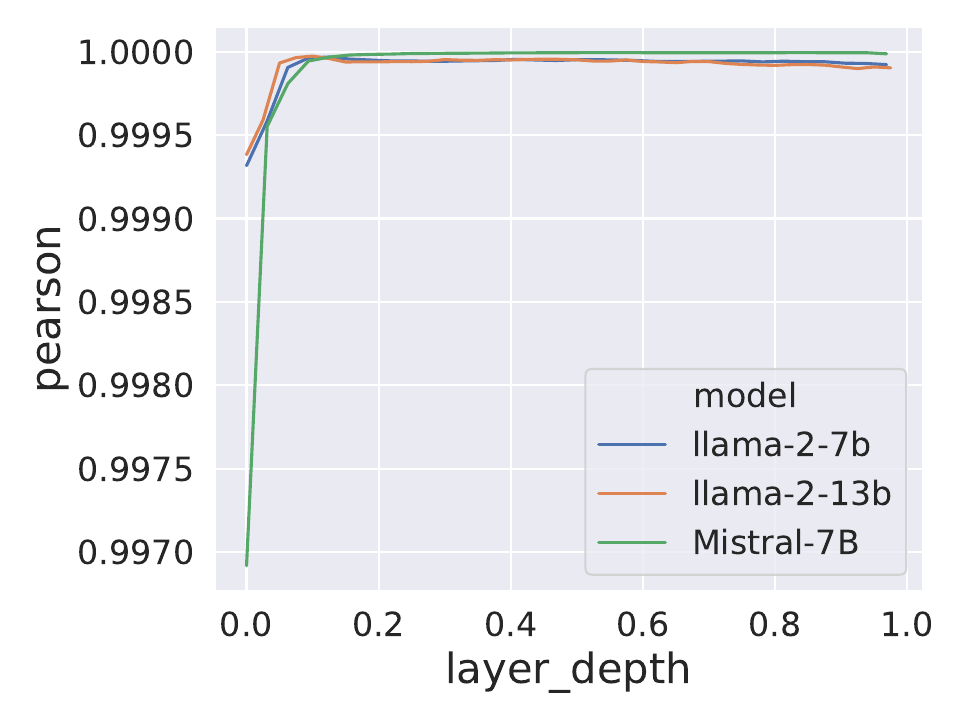}
    }
    \subfloat[$\rho$ of probes on $o$.]{
    \includegraphics[width=0.3\textwidth]{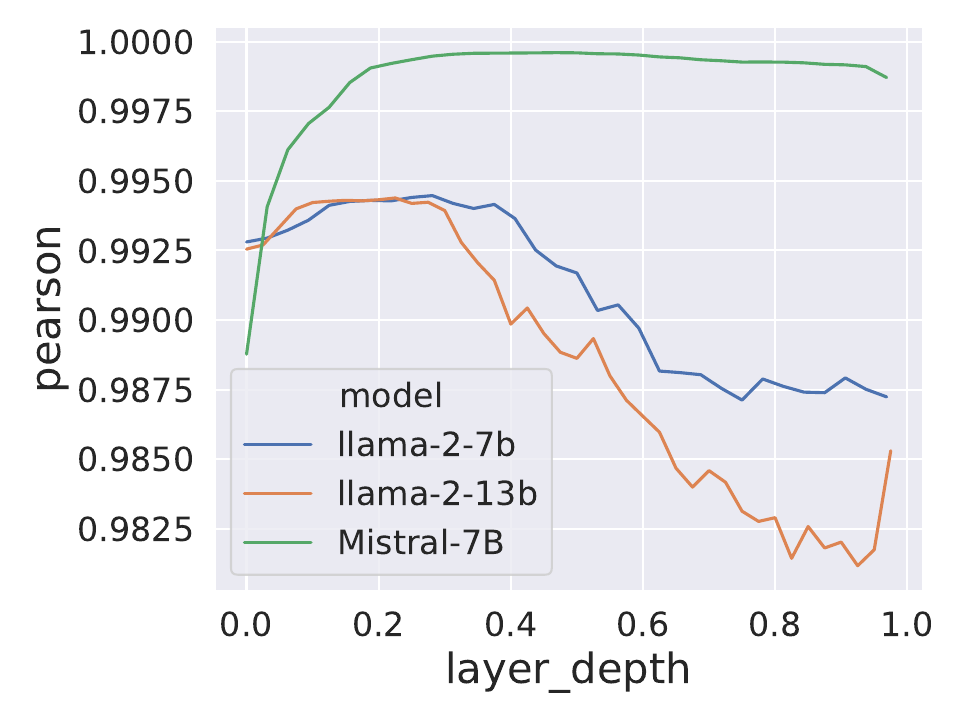}
    } \\
    \subfloat[$R^2$ of probes on $a$.]{
    \includegraphics[width=0.3\textwidth]{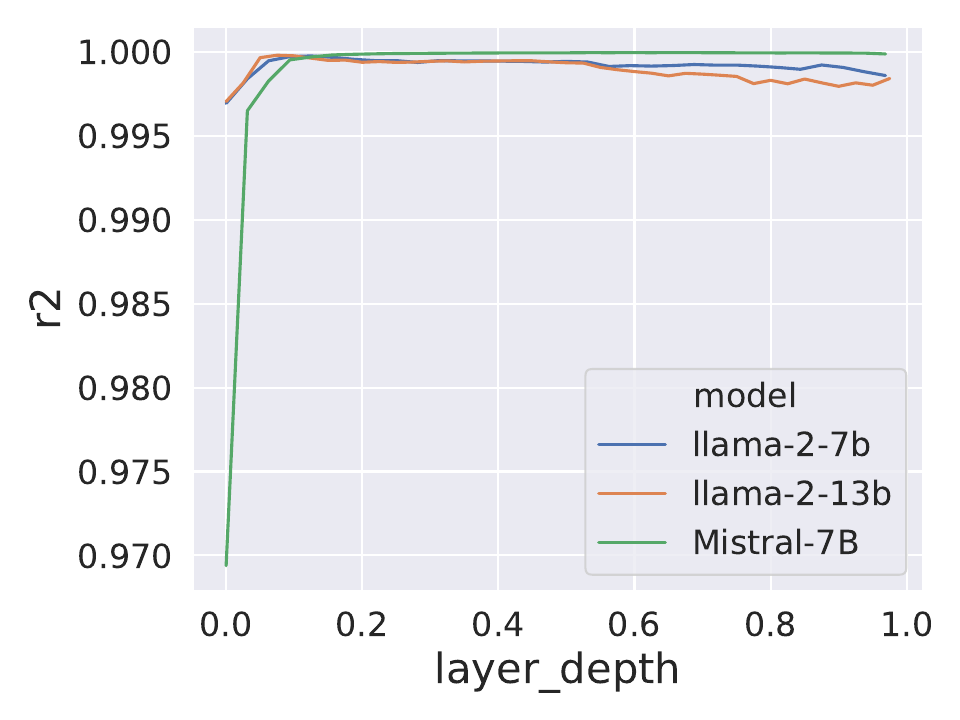}
    }
    \subfloat[$R^2$ of probes on $b$.]{
    \includegraphics[width=0.3\textwidth]{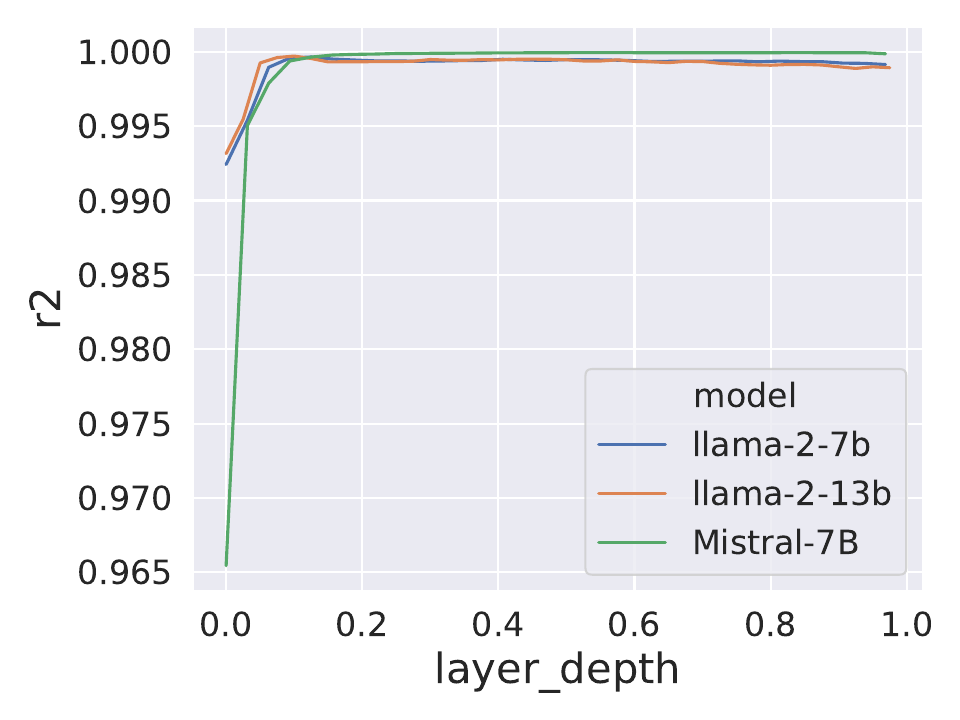}
    }
    \subfloat[$R^2$ of probes on $o$.]{
    \includegraphics[width=0.3\textwidth]{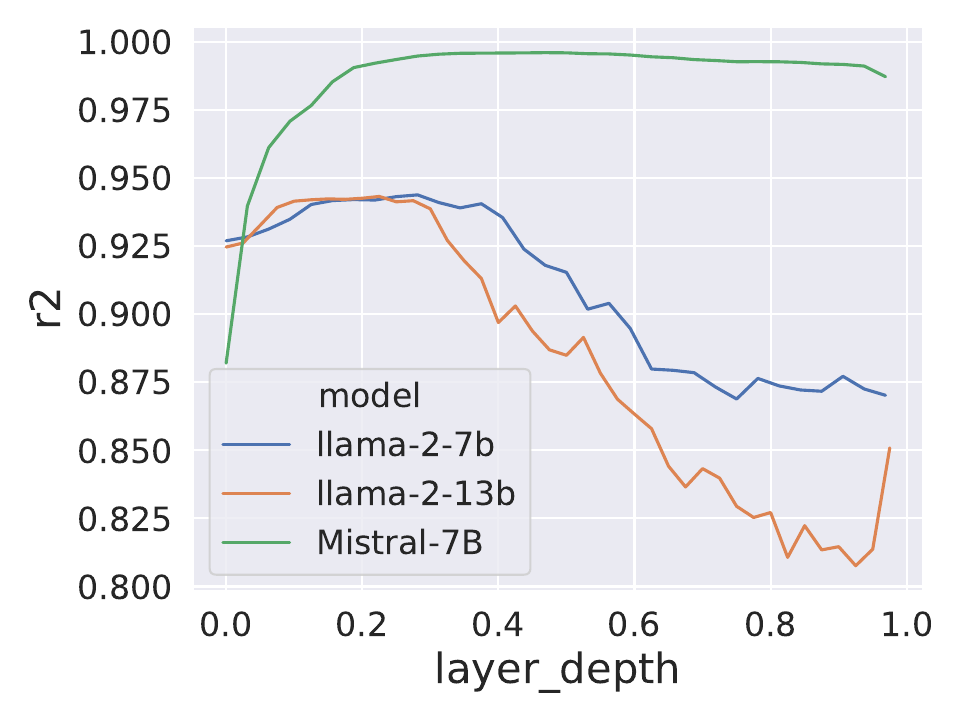}
    } \\
    \subfloat[AAcc of probes on $a$.]{
    \includegraphics[width=0.3\textwidth]{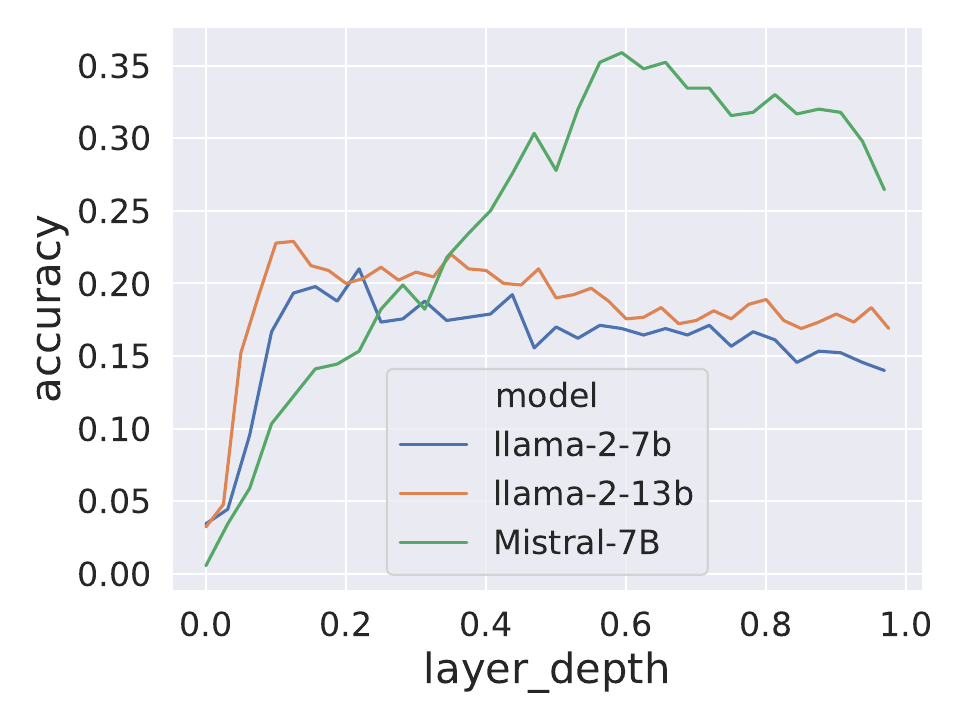}
    }
    \subfloat[AAcc of probes on $b$.]{
    \includegraphics[width=0.3\textwidth]{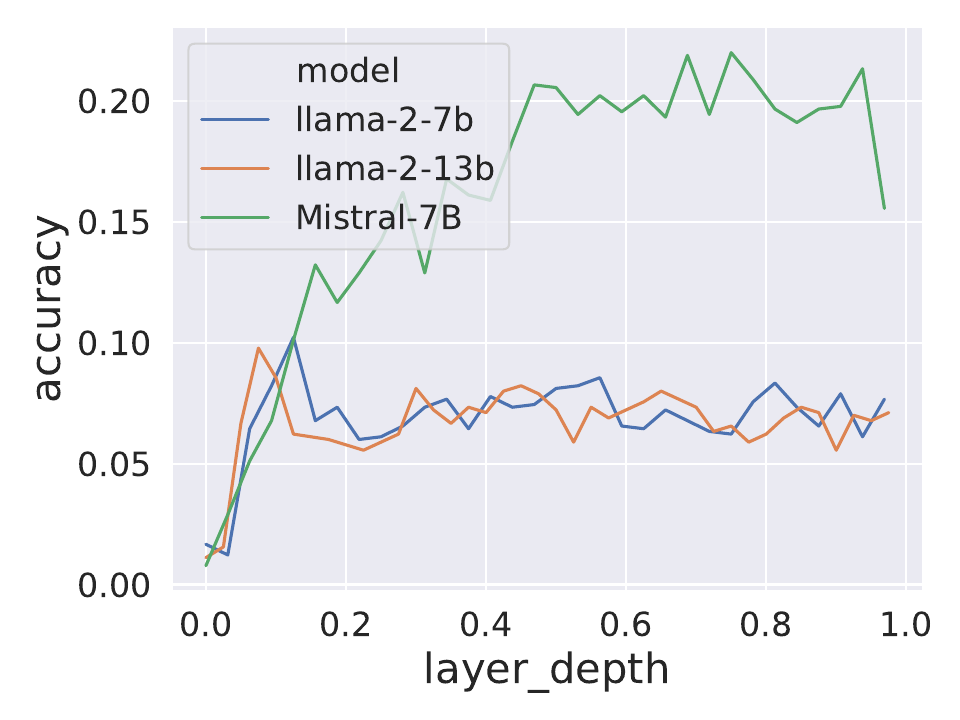}
    }
    \subfloat[AAcc of probes on $o$.]{
    \includegraphics[width=0.3\textwidth]{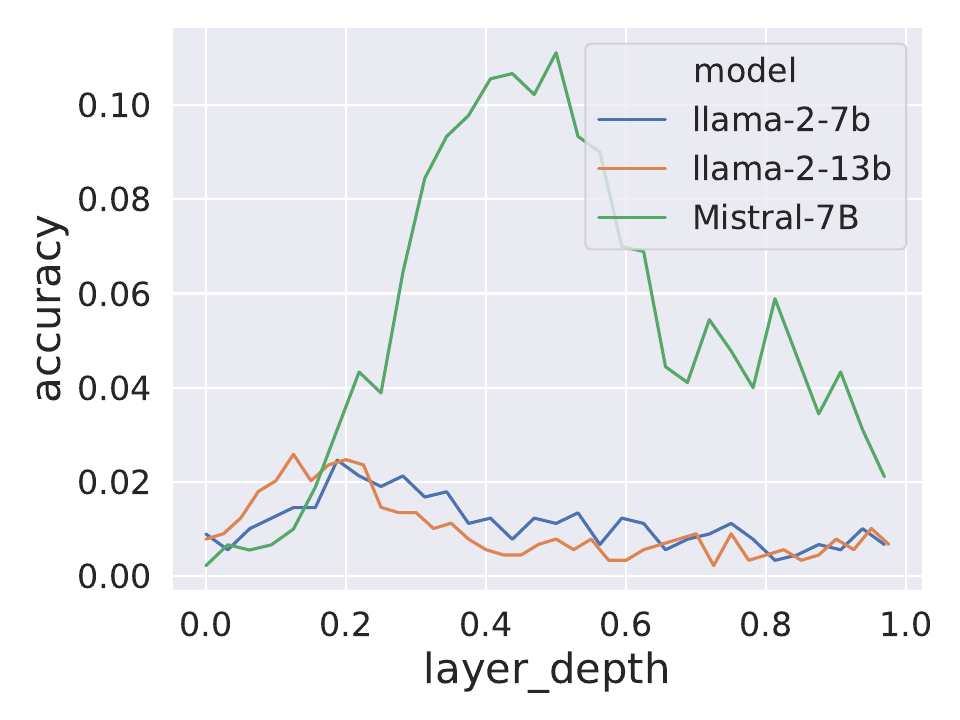}
    } \\
    \subfloat[MSE of probes on $a$.]{
    \includegraphics[width=0.3\textwidth]{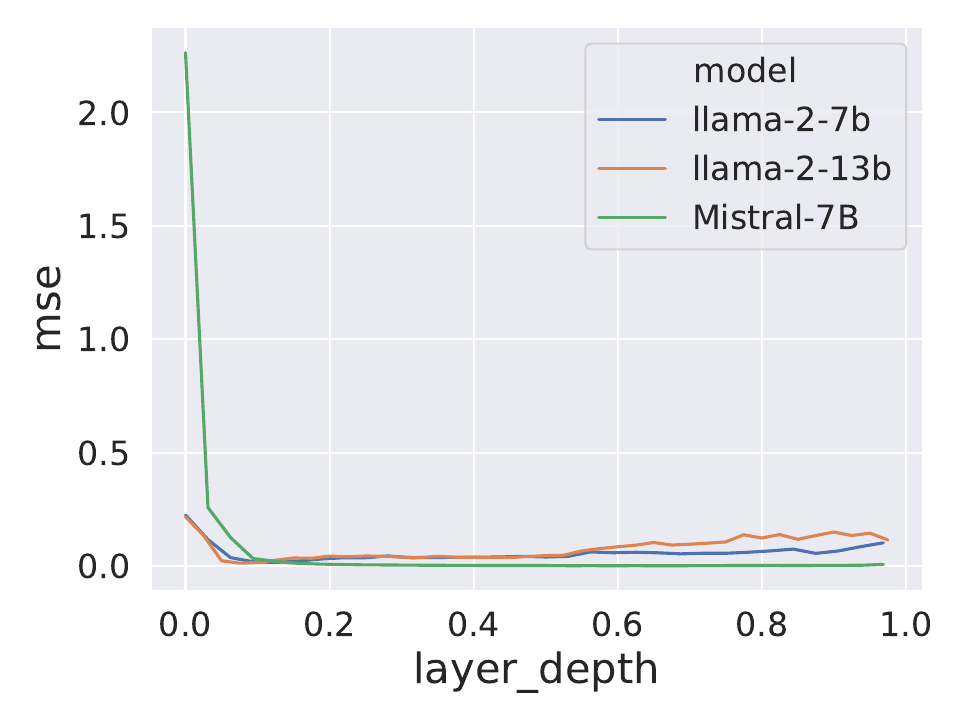}
    }
    \subfloat[MSE of probes on $b$.]{
    \includegraphics[width=0.3\textwidth]{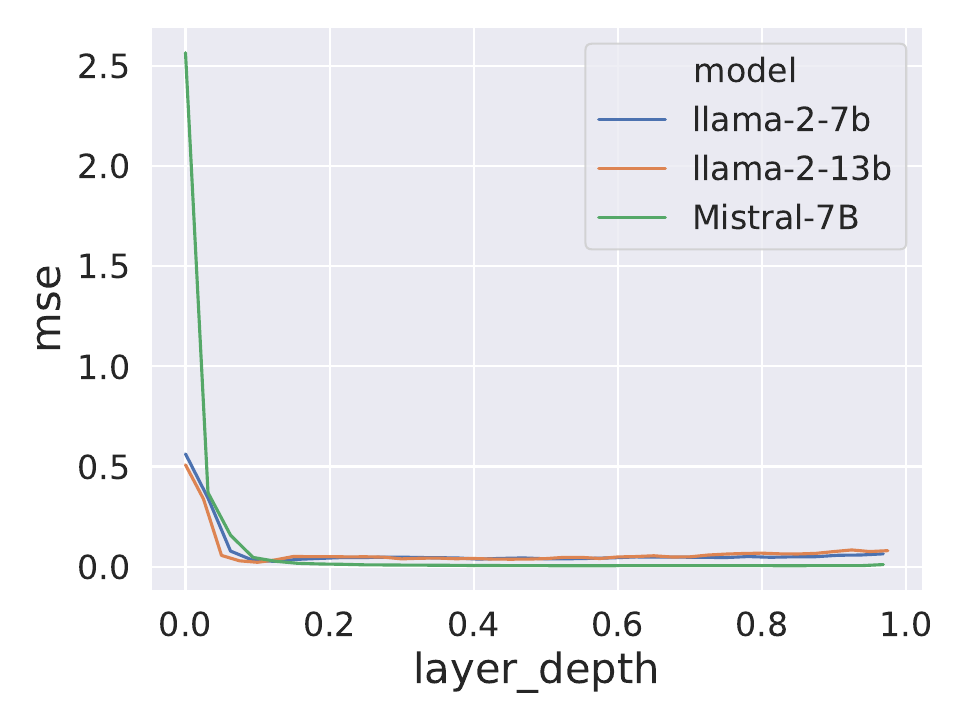}
    }
    \subfloat[MSE of probes on $o$.]{
    \includegraphics[width=0.3\textwidth]{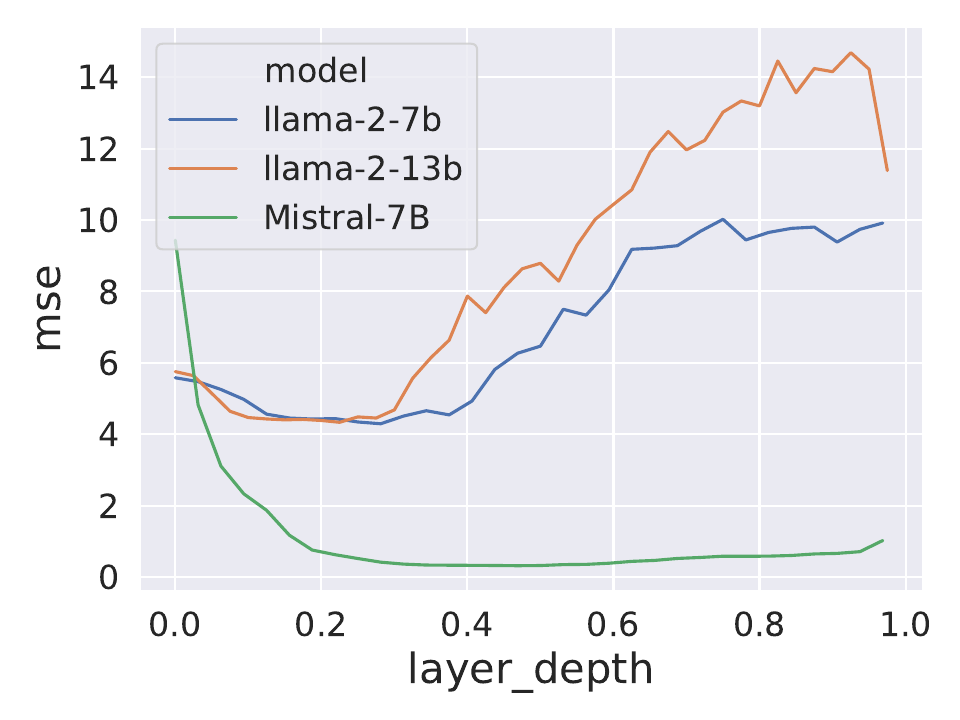}
    }
    \caption{Pearson coefficient ($\rho$), out-of-sample $R^2$, approximate accuracy (AAcc), and mean square error (MSE) of probes on different layers for subtraction problems. $a$ and $b$ refer to the two input numbers denoted in Section \ref{sec:dataset}, and $o$ refers to the prediction of language models respectively. High $\rho$ and $R^2$ indicate the existence of encoded number values in the hidden states.}
    \label{fig:appendix_subtraction_results}
\end{figure*}

\section{Experiments on Subtraction Problems}
\label{sec:appendix_subtract}
In the main paper, we only show the results of probing on addition problems.
We also conduct experiments on subtraction problems with the form of:
\begin{verbatim}
Question: What is the result of {a} 
minus {b}? 
Answer: {a - b}
\end{verbatim}
where we assert $a > b$ to ensure the result being a positive number.

Figure \ref{fig:appendix_subtraction_results} demonstrates the result of probing on subtraction problems.
We can clearly observe that the trends of different metrics are similar to those on addition problems.
In other words, the behaviour of language models on subtraction problems are similar to the behaviour on addition problems.

\section{Overall Accuracy}
\label{sec:appendix_overall_accuracy}
\begin{figure*}[ht]
    \centering
    \subfloat[AAcc of LLaMA-2-7B.]{
    \includegraphics[width=0.3\textwidth]{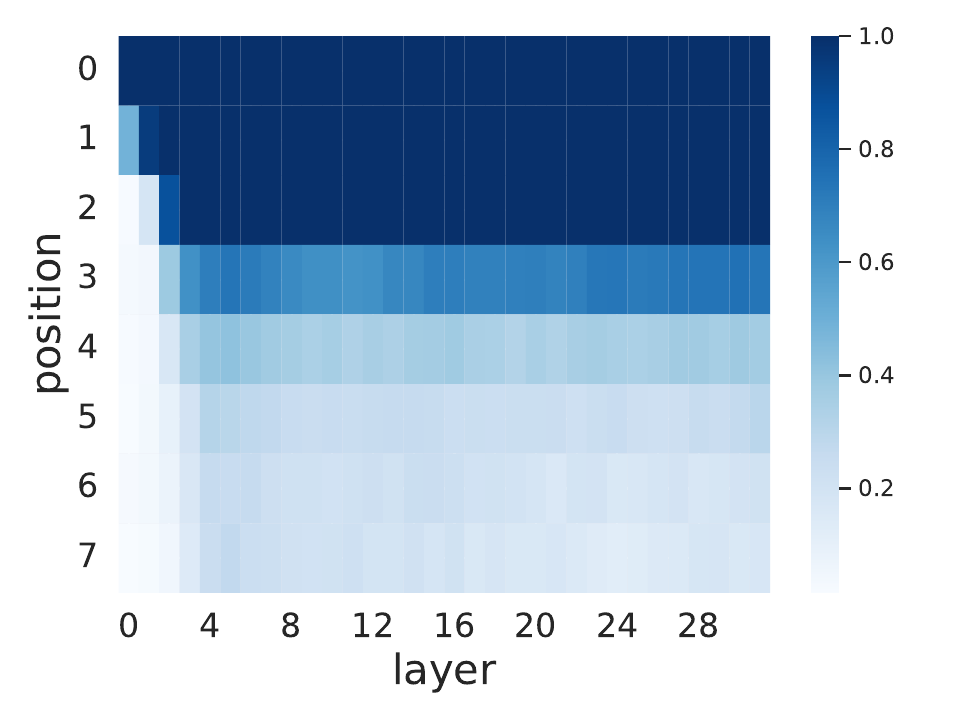}
    }
    \subfloat[AAcc of LLaMA-2-13B.]{
    \includegraphics[width=0.3\textwidth]{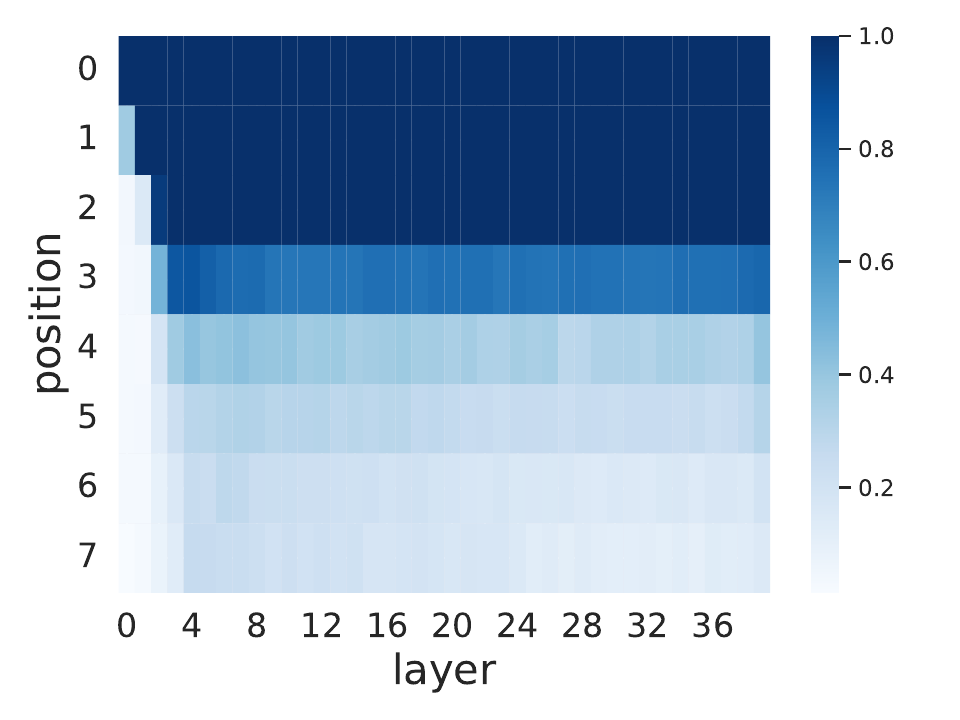}
    }
    \subfloat[AAcc of Mistral-7B.]{
    \includegraphics[width=0.3\textwidth]{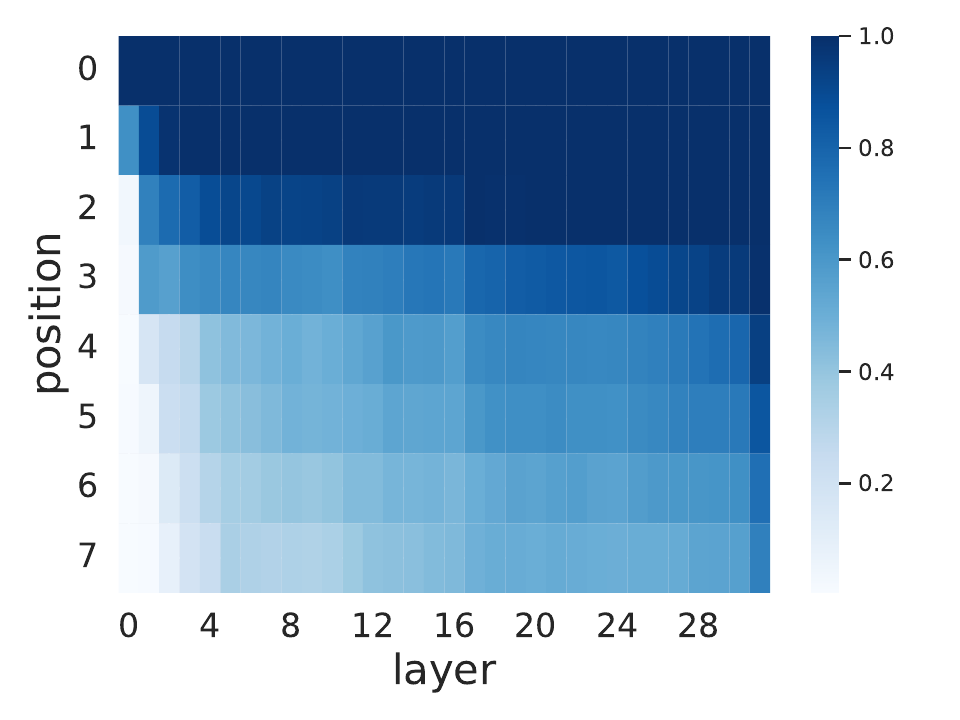}
    }
    \caption{The approximate accuracy (AAcc) of probes on partial number sequence of 8-digit numbers. The y-axis represents the index of number tokens in the token sequence.}
    \label{fig:incremental}
\end{figure*}

Figure \ref{fig:appendix_overall_accuracy} shows the overall accuracy of different language models on addition problems.
We can see that the accuracy of all models, especially LLaMA-2 models, faces a sharp decline at 6-digit problems, which may have a possible correlation with the partial number encoding accuracy demonstrated in Figure \ref{fig:incremental}.

In the LLaMA-2 family, the 13B model does not show any advantage over the 7B model on probing metrics.
In contrast, Mistral-7B displays better performance on all probing metrics, which is consistent with its outstanding math ability.
The difference implies that the ability to encode numbers is consistent across different model scales, but varies between different model families.
Meanwhile, the ability to understand numbers show a positive correlation with the math ability of LLMs.

\begin{figure*}[ht]
    \centering
    \subfloat[Pearson coefficient]{
    \includegraphics[width=0.3\textwidth]{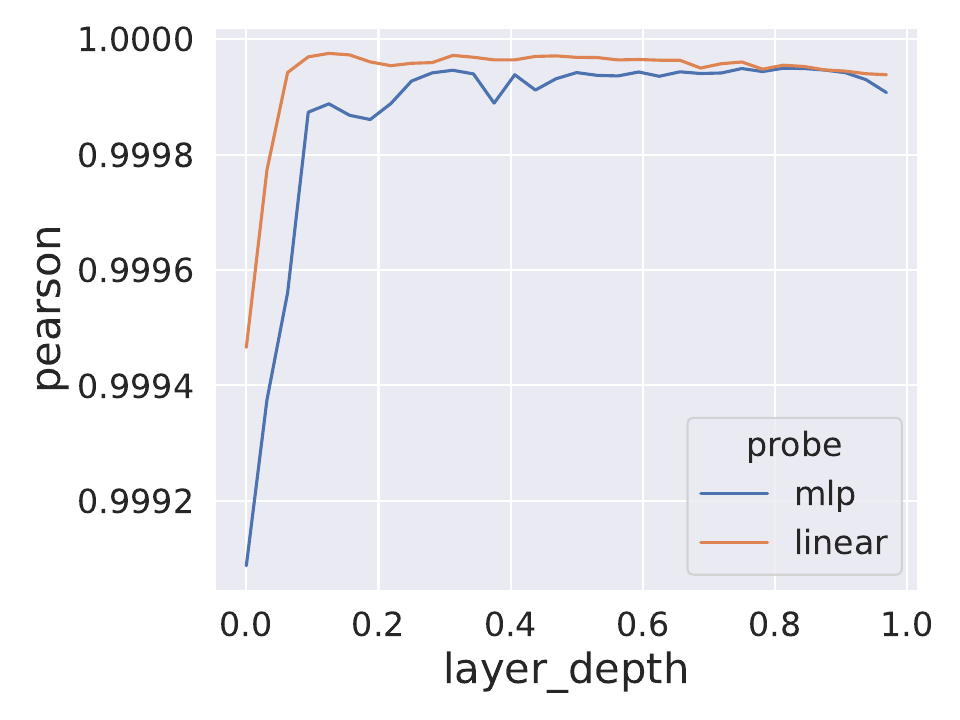}
    }
    \subfloat[Out of sample $R^2$]{
    \includegraphics[width=0.3\textwidth]{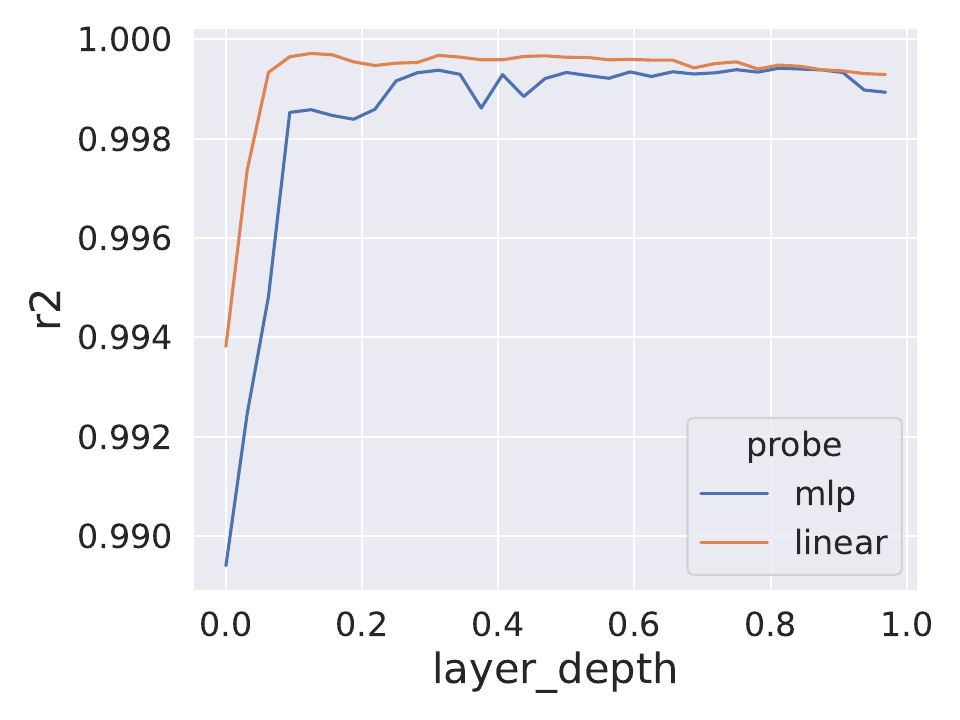}
    }
    \subfloat[Mean square error]{
    \includegraphics[width=0.3\textwidth]{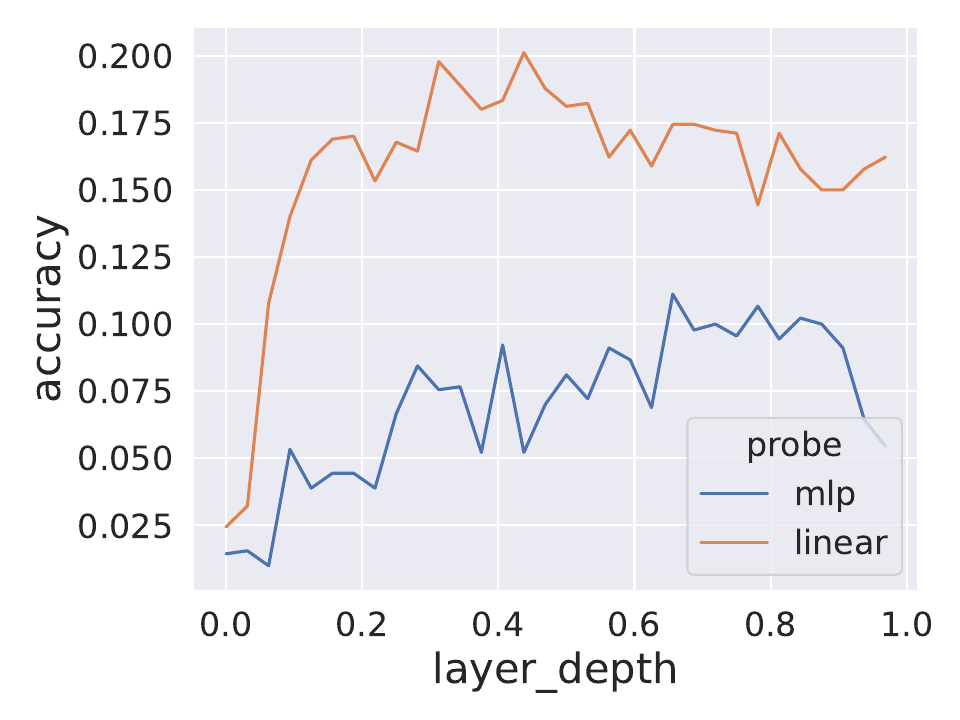}
    }
    \caption{Comparison between linear probes and non-linear MLP probes. Pearson coefficient, out-of-sample $R^2$, and AAcc of probes on the first input number $a$ on different layers are shown in the figure.}
    \label{fig:appendix_linearity}
\end{figure*}

\section{Detailed Experiments on Linearity}
\label{sec:appendix_linearity}
Figure \ref{fig:appendix_linearity} shows the comparison between linear probes and MLP probes on $\rho$, $R^2$ and MSE.
We can observe that MLP probes generally perform no better than linear probes.

\begin{figure*}[ht]
    \centering
    \subfloat[$\rho$ of LLaMA-2-7B.]{
    \includegraphics[width=0.3\textwidth]{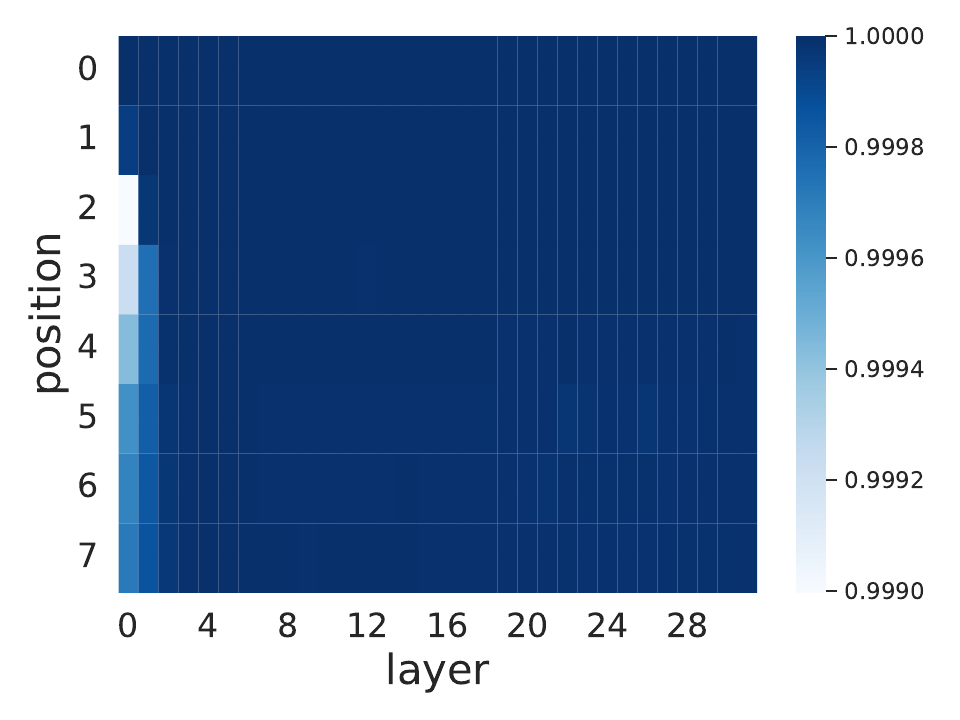}
    }
    \subfloat[$\rho$ of LLaMA-2-13B.]{
    \includegraphics[width=0.3\textwidth]{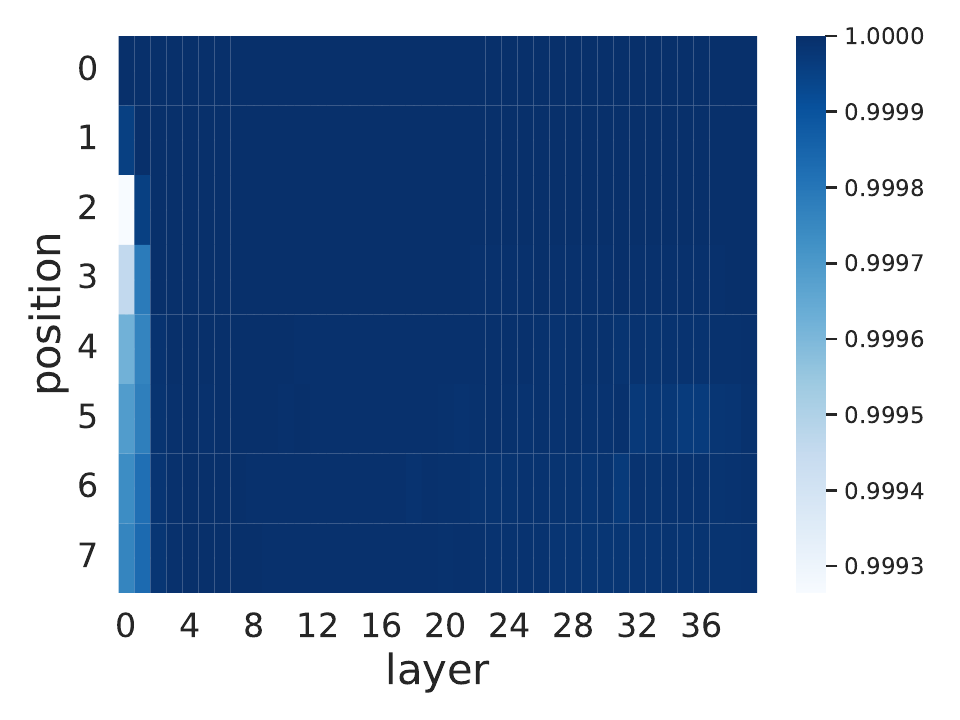}
    }
    \subfloat[$\rho$ of Mistral-7B.]{
    \includegraphics[width=0.3\textwidth]{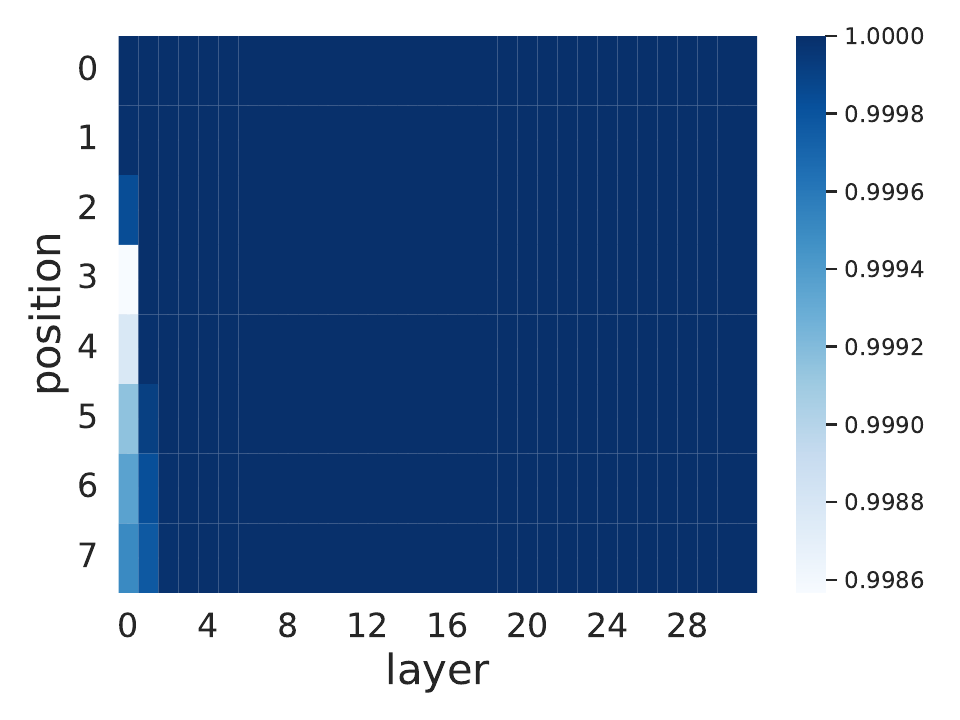}
    }
    \\
    \subfloat[$R^2$ of LLaMA-2-7B.]{
    \includegraphics[width=0.3\textwidth]{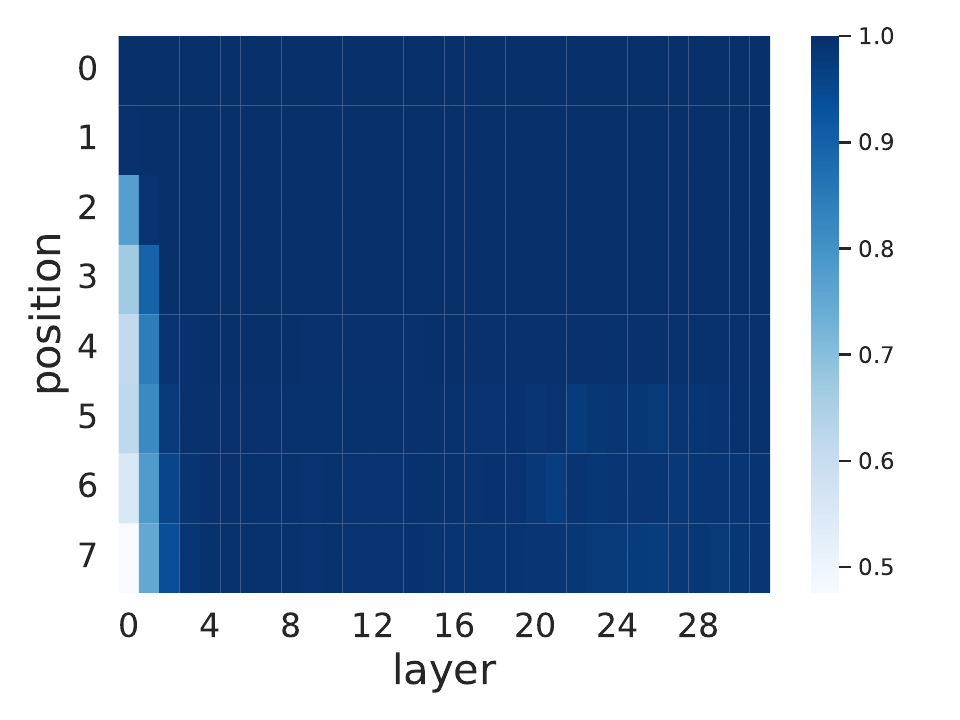}
    }
    \subfloat[$R^2$ of LLaMA-2-13B.]{
    \includegraphics[width=0.3\textwidth]{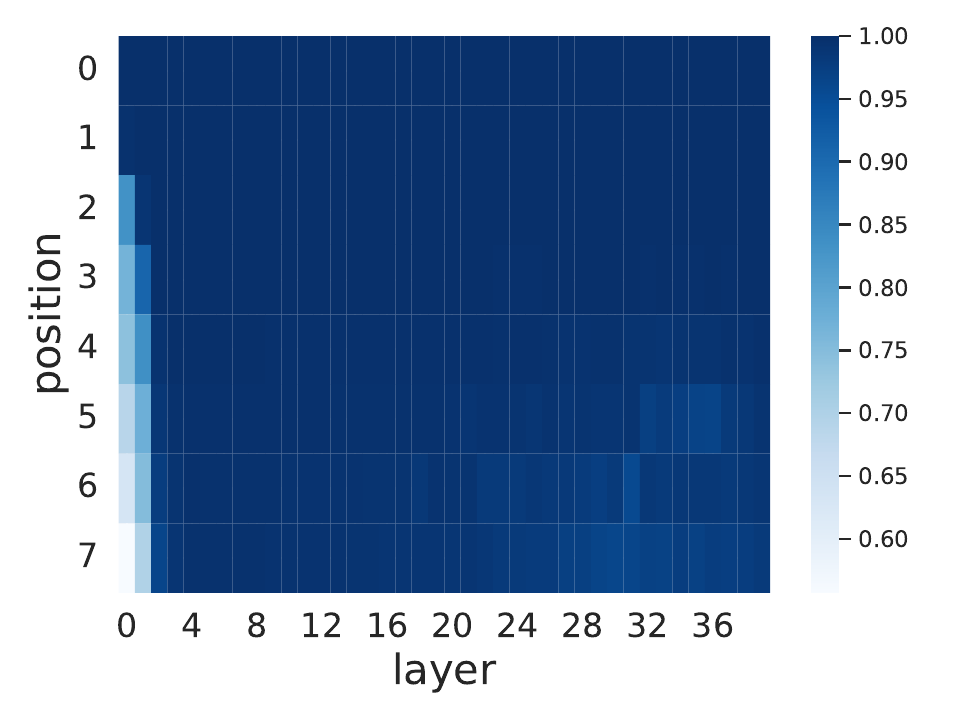}
    }
    \subfloat[$R^2$ of Mistral-7B.]{
    \includegraphics[width=0.3\textwidth]{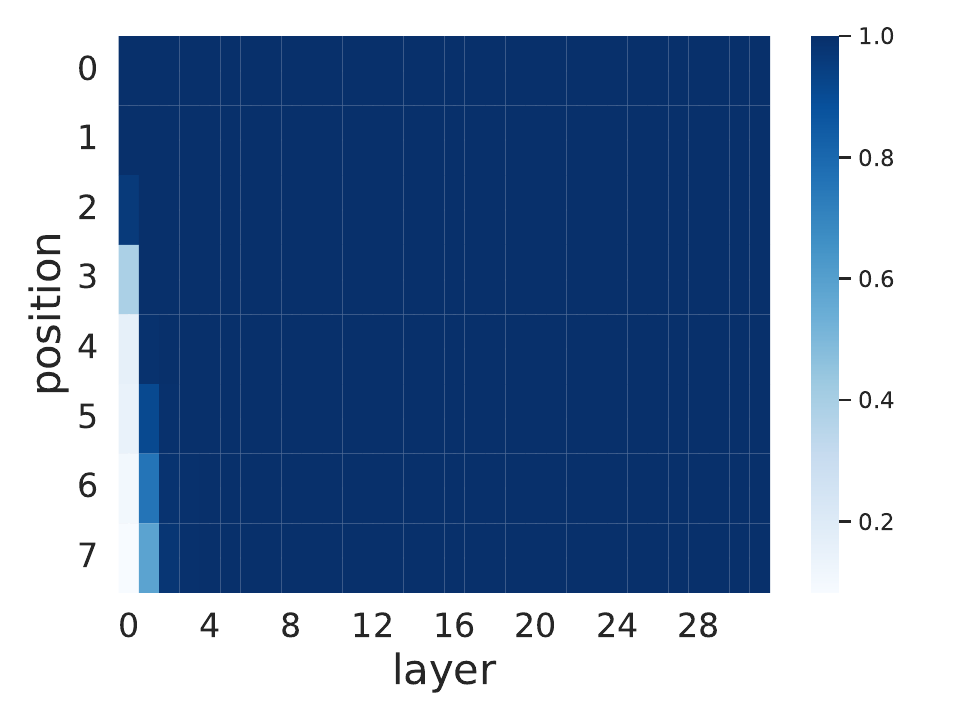}
    }
    \\
    \subfloat[MSE of LLaMA-2-7B.]{
    \includegraphics[width=0.3\textwidth]{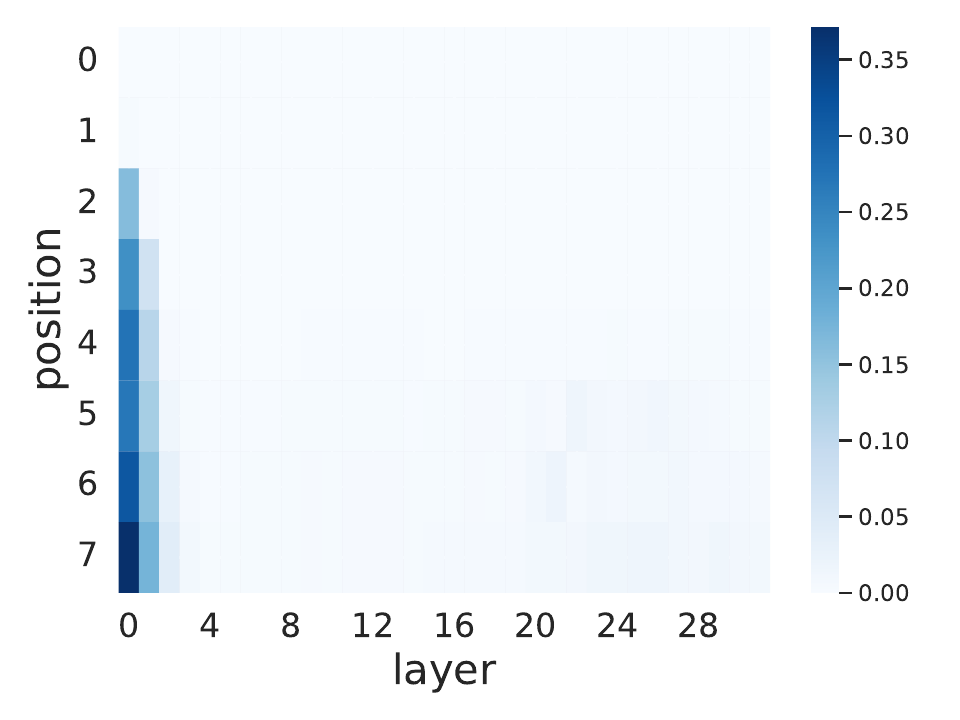}
    }
    \subfloat[MSE of LLaMA-2-13B.]{
    \includegraphics[width=0.3\textwidth]{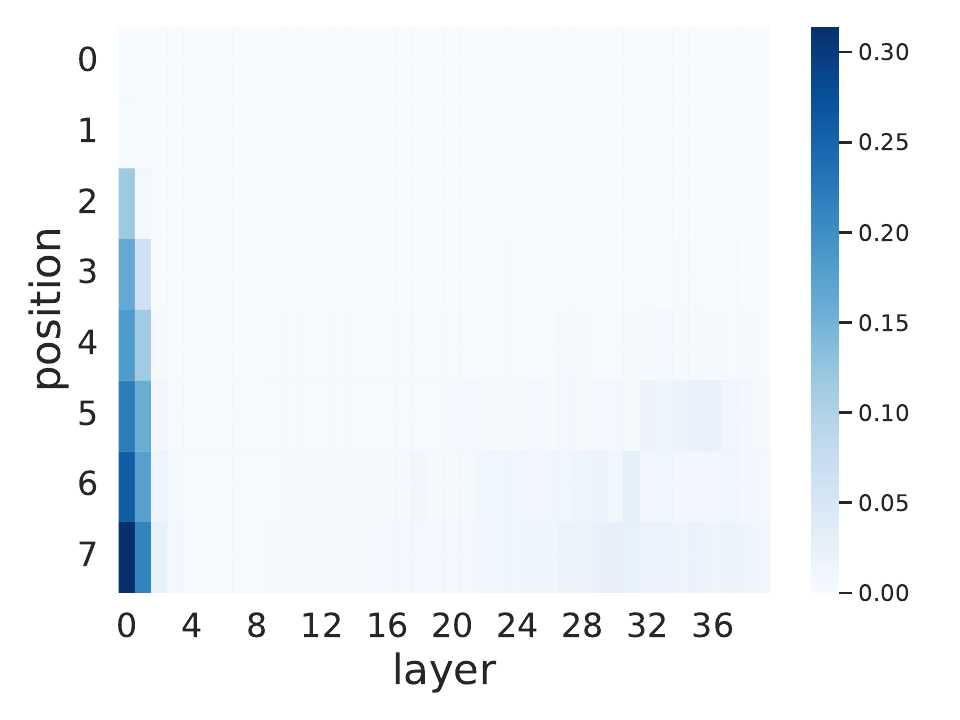}
    }
    \subfloat[MSE of Mistral-7B.]{
    \includegraphics[width=0.3\textwidth]{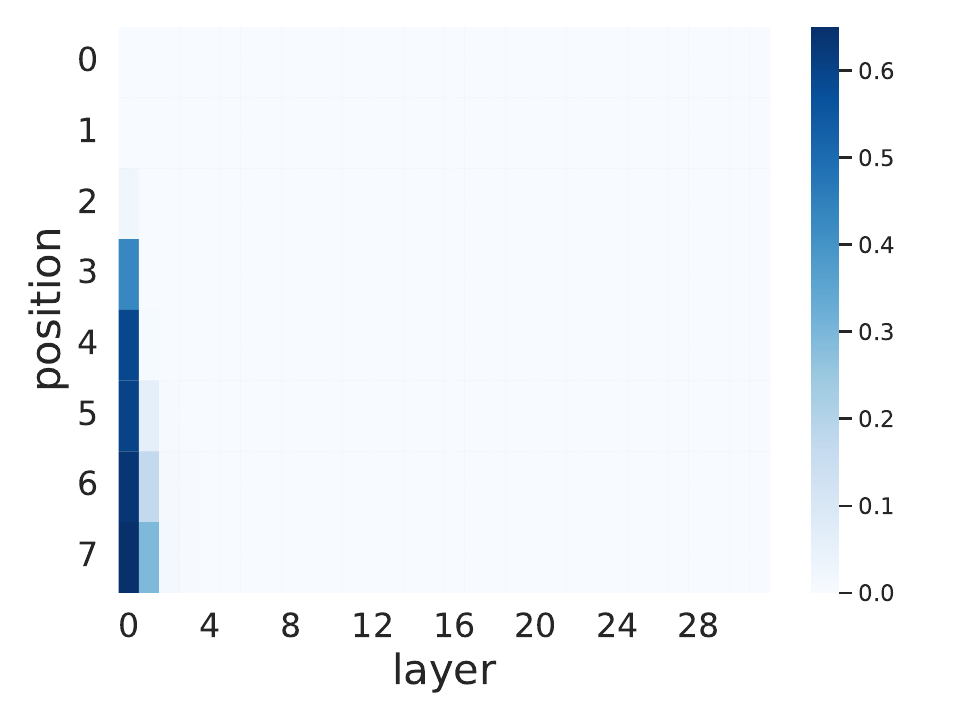}
    }
    \caption{The Pearson coefficient ($\rho$), out-of-sample $R^2$, and mean square error (MSE) of probes on partial number sequence of 8-digit numbers. The y-axis represents the index of number tokens in the token sequence.}
    \label{fig:appendix_incremental}
\end{figure*}

\section{Experimental Results on Partial Number Encoding}
\label{sec:appendix_partial}
In large language models like LLaMA-2, large numbers are split into multiple tokens, where each token represents a certain digit of the original number.
This raises a question: whether the encoding process will proceed from token to token, or will it only happen at the end of number token sequences?

To investigate the problem, we choose addition problems consisting of 8-digit numbers and probe the value of the partial number sequence at every token position.
For example, given a number token sequence ``12345678'', we will probe the value 12 at the position of token ``2'', and probe the value 123 at the position of token ``3''.

Figure \ref{fig:incremental} shows the probing accuracy of 3 models.
It can be observed that the value of the partial number sequence can be read out at every token position. In other words, language models encode the number token sequence incrementally.

Meanwhile, the accuracy significantly declines as the token sequence gets longer, which means that language models face increasing difficulty in capturing the precise value as the number gets larger in scale.
Notice that Mistral-7B suffers less from accuracy decay, we can assume that the ability to precisely encode long number token sequences is positively correlated to the mathematical ability of language models.

Figure \ref{fig:appendix_incremental} shows the Pearson coefficient, out-of-sample $R^2$, and mean square error of probes on partial sequence of 8-digit numbers.
These metrics remain stable as the length of number token sequence gets longer, indicating that language models do have the ability to incrementally encode number values, but there would be more error when the number gets larger in scale.

\begin{figure*}[ht]
    \centering
    \subfloat[$a$ versus $c$ of LLaMA-2-7B.]{
    \includegraphics[width=0.3\textwidth]{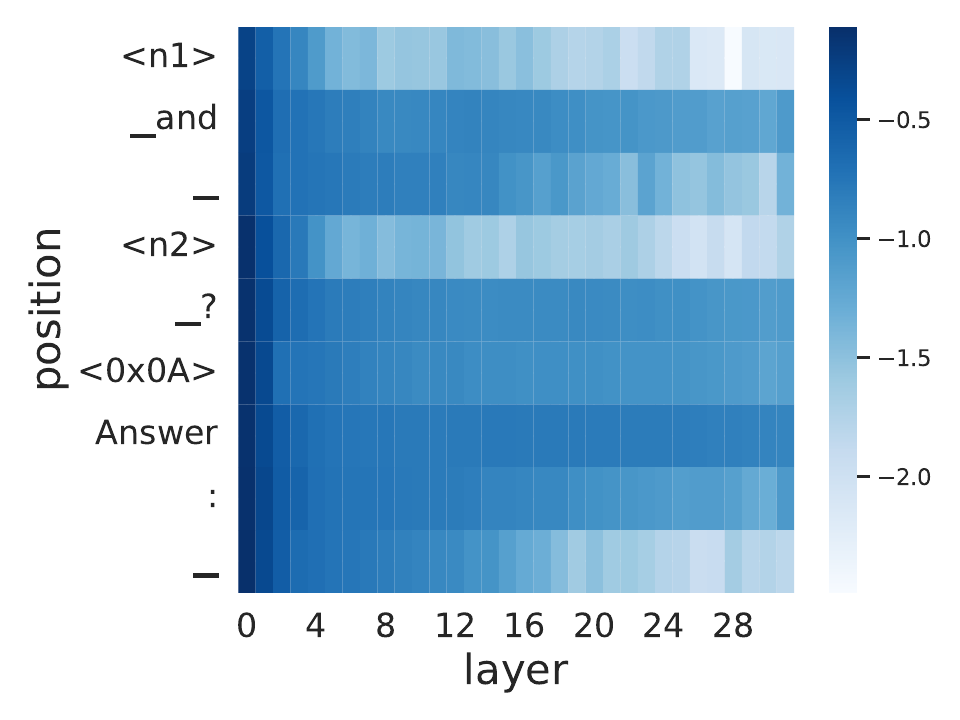}
    }
    \subfloat[$a$ versus $c$ of LLaMA-2-13B.]{
    \includegraphics[width=0.3\textwidth]{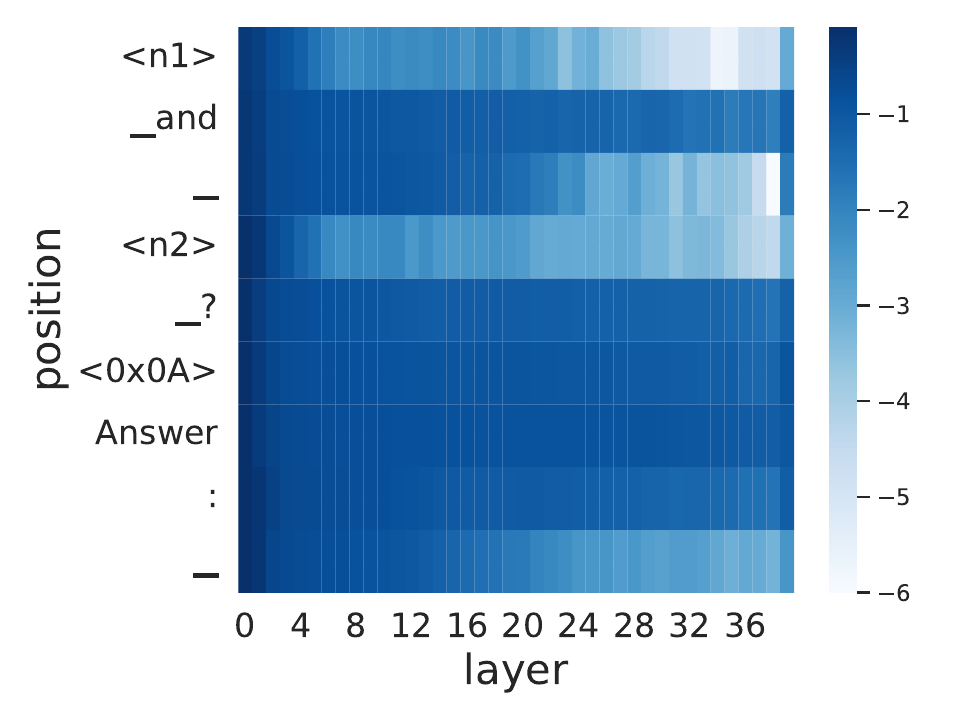}
    }
    \subfloat[$a$ versus $c$ of Mistral-7B.]{
    \includegraphics[width=0.3\textwidth]{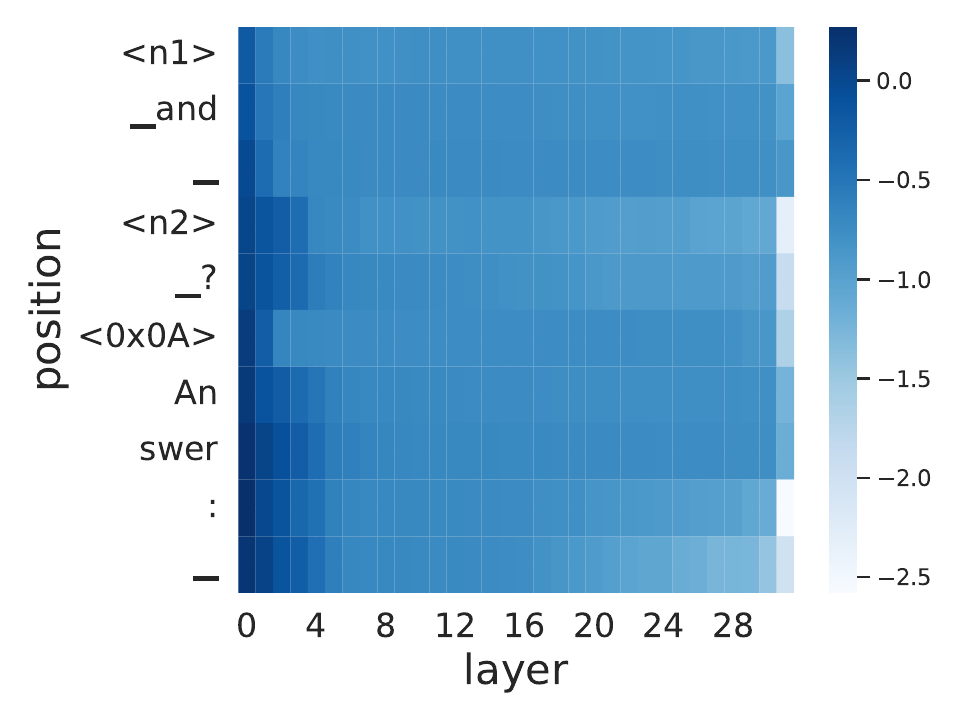}
    }
    \\
    \subfloat[$b$ versus $c$ of LLaMA-2-7B.]{
    \includegraphics[width=0.3\textwidth]{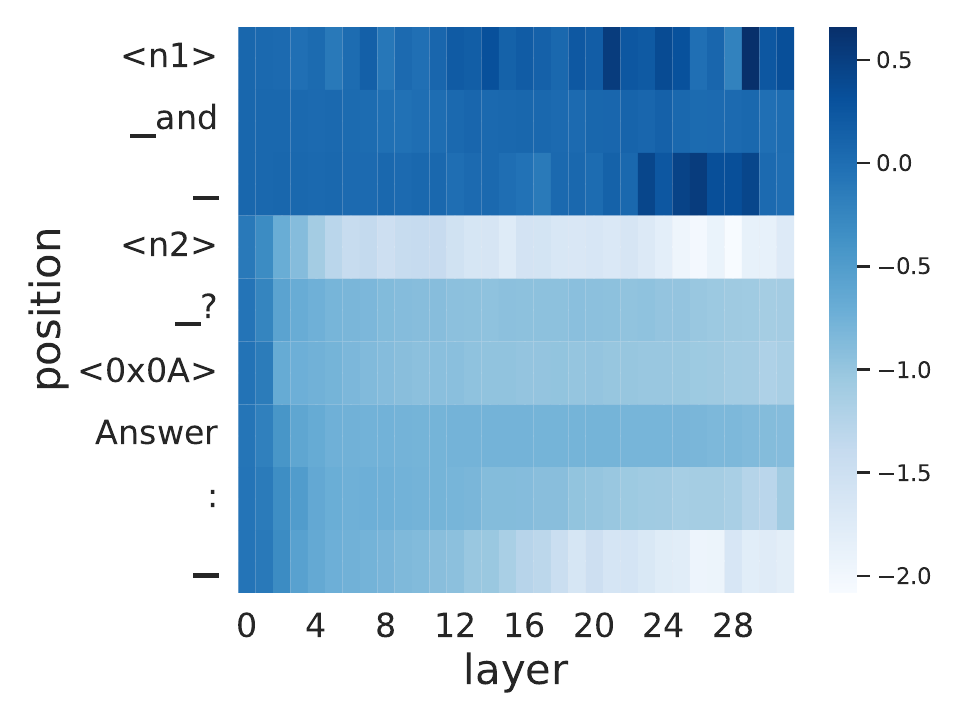}
    }
    \subfloat[$b$ versus $c$ of LLaMA-2-13B.]{
    \includegraphics[width=0.3\textwidth]{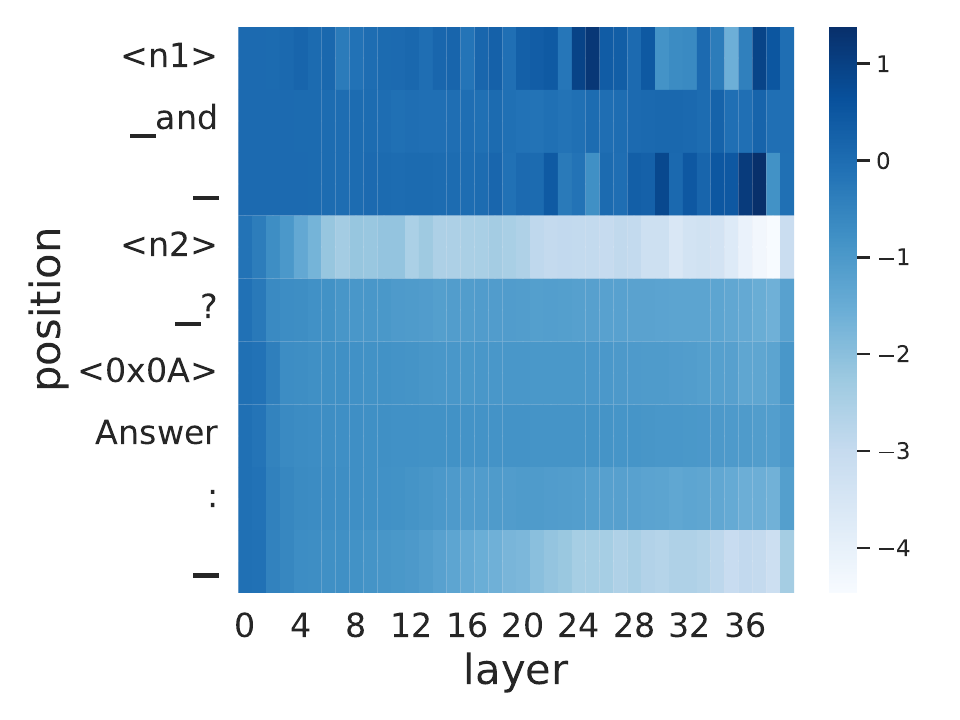}
    }
    \subfloat[$b$ versus $c$ of Mistral-7B.]{
    \includegraphics[width=0.3\textwidth]{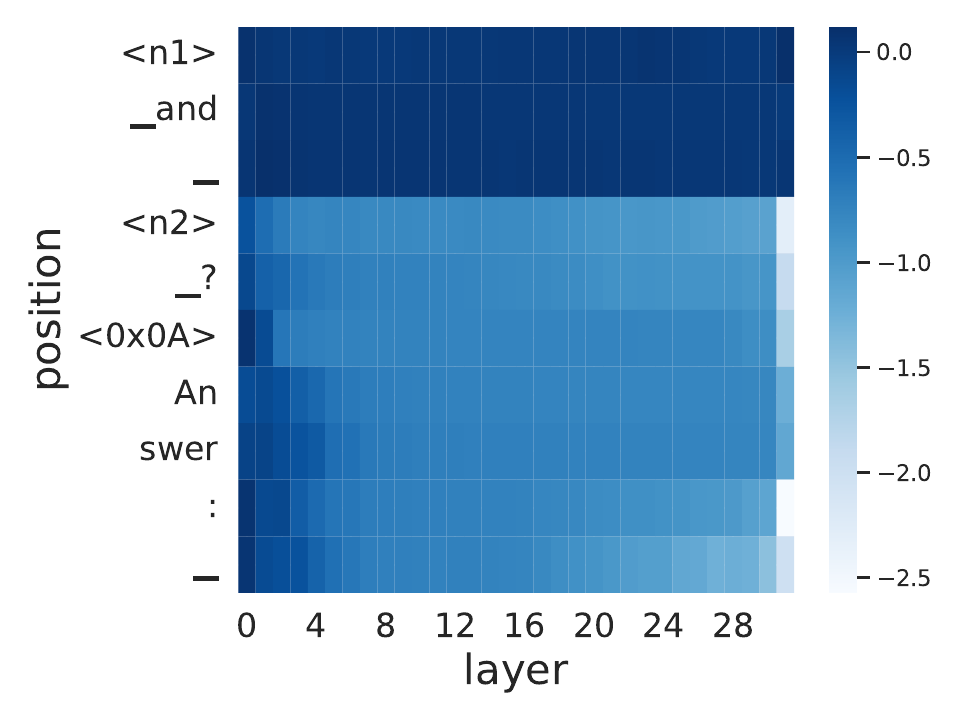}
    }
    \caption{The difference in mean square error (MSE) between probes on input numbers and control signals. A lighter color indicates a greater performance gap.}
    \label{fig:appendix_control}
\end{figure*}

\section{Probing With Control Tasks}
There exists the risk that probes may learn to extract values that language models do not encode.
In Figure \ref{fig:persistence}, we can see that probing on the second input number $b$ at positions before it appears would lead to extremely large mean square errors, which acts as a piece of preliminary evidence that the probe performance does not solely come from probe strength.

To quantify the influence of probe strength, we conduct an experiment that probes with control tasks.
For each question, we generate a random number $c$ that shares the same digit with $a$ and $b$ as the control signal.
If the probing performance comes from the encoded number values rather than probe strength, there would be a clear gap between the probing performance on $c$ and $a, b$.

Figure \ref{fig:appendix_control} shows the difference between probe performances.
It can be observed that probing on input numbers constantly yields better performance than probing on random control signals, proving that language models do encode number values in their hidden states.

Meanwhile, probing $b$ on positions before $b$ shows performance similar to probing $c$, which corresponds to the fact that $b$ is unknown to the model at these positions.

\begin{table}[htbp]
    \centering
    \begin{tabular}{c|c|p{0.4\linewidth}}
    \hline
    Patching & Result & Explanation \\
    \hline
    None & 6912 & 5678+1234=6912 \\
    Full & 11233 & 9999+1234=11233 \\
    $5 \xrightarrow{} 9$ & 10912 & 9678+1234=10912 \\
    $6 \xrightarrow{} 9$ & 7212 & 5978+1234=7212 \\
    $7 \xrightarrow{} 9$ & 6932 & 5698+1234=6932 \\
    $8 \xrightarrow{} 9$ & 6913 & 5679+1234=6913 \\
    \hline
    \end{tabular}
    \caption{Patching results on the question ``Question: What is the sum of 5678 and 1234 ?'' by patching the activation on layer 8.}
    \label{tab:appendix_patch_example}
\end{table}

\section{Detailed Experiments on Activation Patching}
\label{sec:appendix_patch}
Table \ref{tab:appendix_linear_example} shows the results of patching on layer 8 of Mistral-7B on the question ``Question: What is the sum of 5678 and 1234 ?''

We can clearly see that patching a digit will only influence the value of the digit itself, rather than the value of the partial token sequence: patching the last digit 8 in 5678 equals changing the number to 5679 rather than 9999, although the encoded value of 9999 can be found in the activation.
We hypothesize that language models encode the number values from scratch at every new position, rather than using previous encoded values.

We also notice that patching the last number digit on early layers shows a higher effect than expected, but the reason why the last digit is more special is still unknown.

\begin{figure}[ht]
    \centering
    \includegraphics[width=0.45\textwidth]{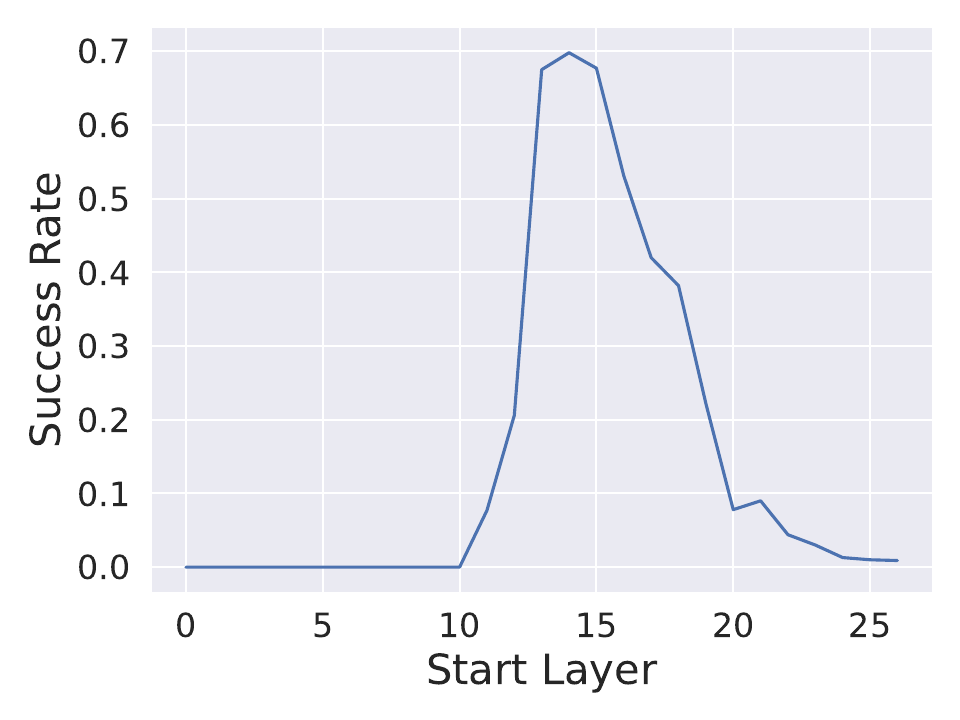}
    \caption{The success rate of performing a linear intervention on 5 consecutive layers.}
    \label{fig:appendix_linear_5}
\end{figure}

\begin{figure}[ht]
    \centering
    \includegraphics[width=0.45\textwidth]{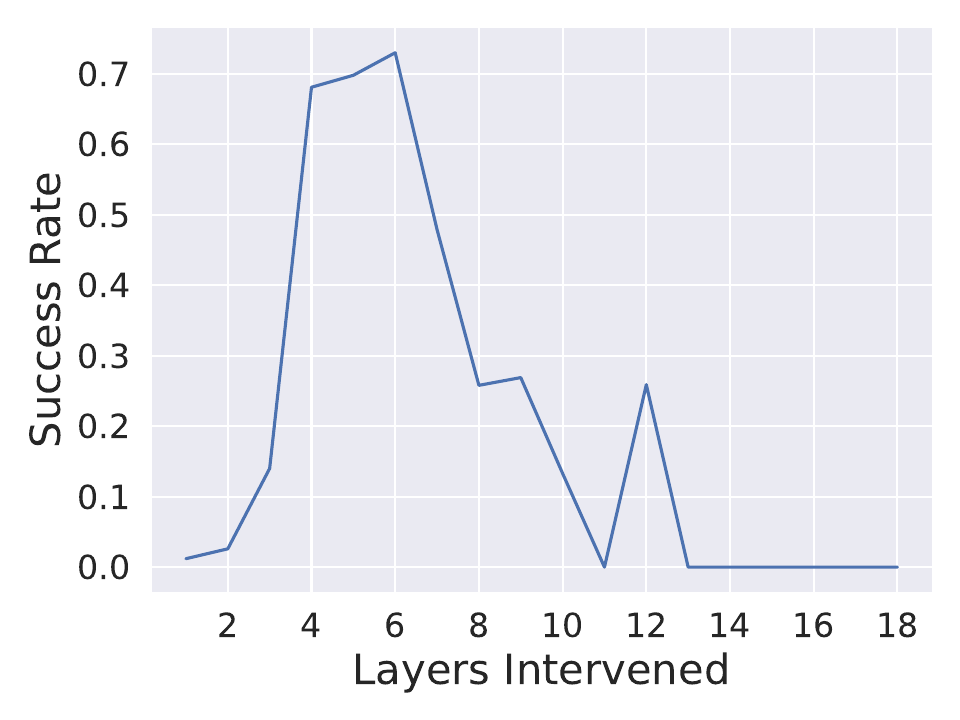}
    \caption{The success rate of performing a linear intervention on layers starting from layer 14.}
    \label{fig:appendix_linear_14}
\end{figure}

\section{Detailed Experiments on Linear Intervention}
\label{sec:appendix_linear}
\subsection{Success Rate}
Figure \ref{fig:appendix_linear_5} shows the success rate of intervening on 5 consecutive layers with a maximum success rate of 0.698, and Figure \ref{fig:appendix_linear_14} shows the success rate of intervening on a series of layers starting from layer 14. 
It can be observed that a sufficient number of layers need to be intervened for language models to successfully change their predictions.
\citet{nanda2023emergent} observed a similar phenomenon in OthelloGPT, and a related hypothesis is that language models demonstrate the Hydra effect~\cite{mcgrath2023hydra}, where other layers would self-repair the intervention on certain layers.

\begin{table}[htbp]
    \centering
    \begin{tabular}{c|p{0.4\linewidth}}
    \hline
    Layer & Generation Result \\
    \hline
    0-5 & Answer:  gainedcnt I I I I I I I I I I C C C \\
    14-19 & Answer: 12515 \\
    25-30 & Answer: 6455 \\
    \hline
    \end{tabular}
    \caption{Intervention results on the question ``Question: What is the sum of 2936 and 3519 ?''. Running Mistral-7B without intervention would lead to the result of 6455.}
    \label{tab:appendix_linear_example}
\end{table}

\subsection{Output Patterns}
We also observe that while intervening on early or late layers both lead to poor success rates, they display different patterns of output.
Table \ref{tab:appendix_linear_example} shows the result of intervening on different layers of Mistral-7B. 
It can be seen that performing a linear intervention on early layers would completely destroy the final outcome, while intervening on late layers will not change the result at all.
We hypothesize that the number encoding in early layers has not fully developed yet, and intervening in it would lead to unexpected results; In late layers, the number encoding is simply remembered but not used, and the language models rely on other subspace to decode the final outcome.

\subsection{Additional Experiments}
We have also tried to change the probed number from the original value $o$ to a new value $o+o'$:
\begin{align}
    \mathbf{h}_{i}\mathbf{W}_{i}+b_{i} &= o \\
    \mathbf{d}_{i} &= o'\frac{\mathbf{W}_{i}}{|\mathbf{W}_{i}|^{2}} \\
    (\mathbf{h}_{i}+\mathbf{d}_{i})\mathbf{W}_{i}+b_{i} &= o+o'
\end{align}
However, the intervention does not yield results as expected: the intervened model continues to predict $o$ rather than $o+o'$.  

A possible hypothesis is that the probed number value is the projection of $\mathbf{h}_{i}$ along the direction $\mathbf{W}_{i}$, and simply adding vectors to $\mathbf{h}_{i}$ would draw it away from its valid subspace.
To maintain intervened $\mathbf{h}_{i}$ in its valid subspace, it should be rotated along certain direction.
The method of precisely changing the encoded number values in language models still remains to be explored.

\begin{figure}[htbp]
    \centering
    \includegraphics[width=0.42\textwidth]{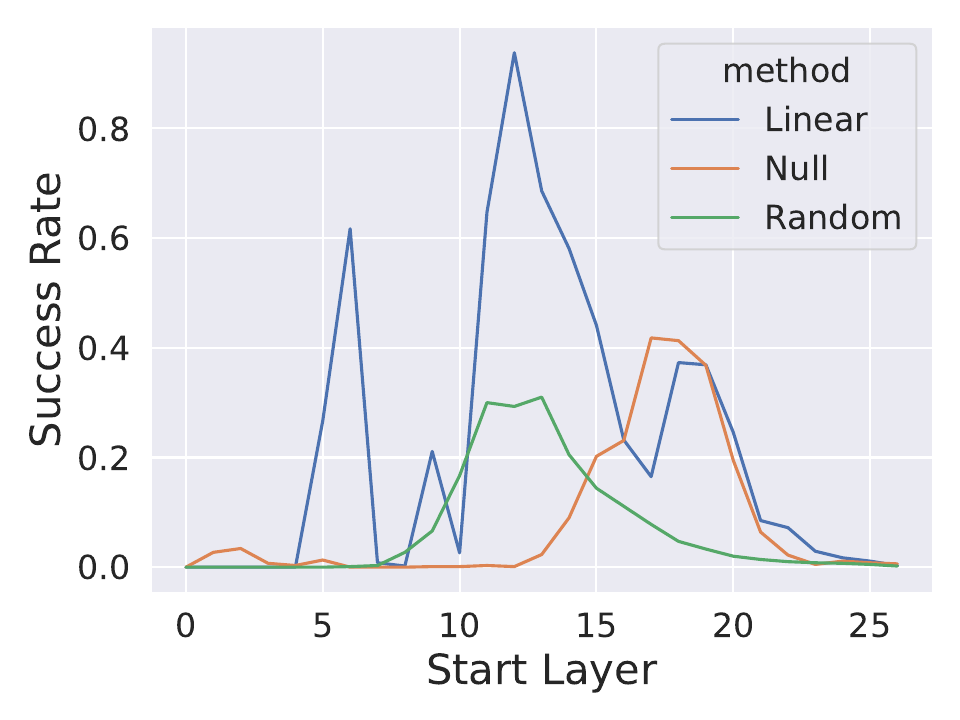}
    \caption{The success rate of performing a linear intervention on 6 consecutive layers, with a negative $\alpha = -2.0$}
    \label{fig:appendix_negative_intervention}
\end{figure}

We also experimented on negative $\alpha$ values, which will "push" the residual stream towards a smaller encoded number.
The results are demonstrated in Figure \ref{fig:appendix_negative_intervention}.
We can see that the trend of success rate is similar to the trend in Figure \ref{fig:intervene_linear}, further proving that the value of calculation result can be linearly intervened.

\begin{figure*}
    \centering
    \subfloat[logMSE]{
    \includegraphics[width=0.4\textwidth]{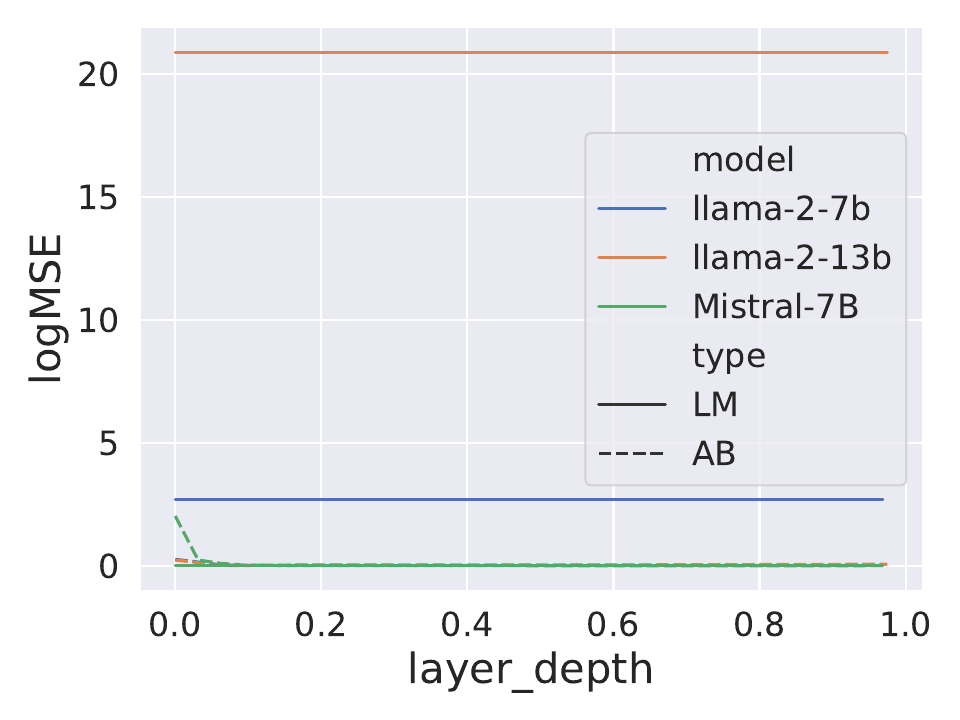}
    }
    \subfloat[Error margin]{
    \includegraphics[width=0.4\textwidth]{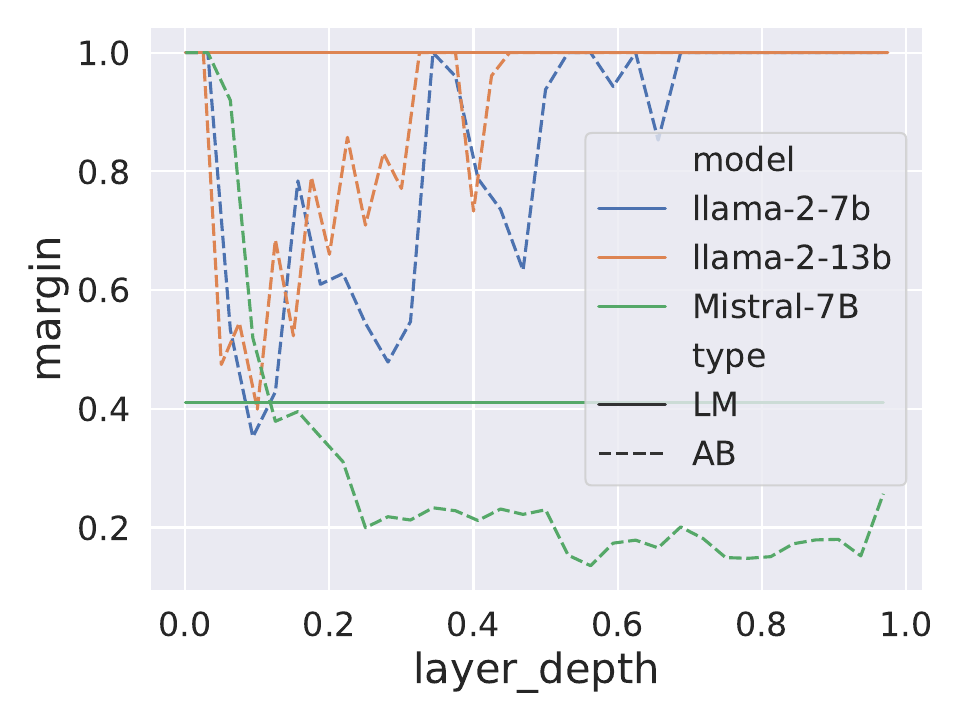}
    }
    \caption{Comparison between the sum of probed $(a, b)$ and language model predictions. AB means the sum of probed $(a, b)$ and LM means language model predictions.}
    \label{fig:addition_comparision}
\end{figure*}

\section{Directly Calculate with Encoded Number Values}
\label{sec:appendix_direct}
We are curious about whether the probed number values could help LLMs better perform calculations.
Considering that adding the probed input numbers does not yield precise answers (Section \ref{ssec:number_existence}), we evaluate the sum of probed numbers with two new metrics: logMSE and error margin.

\begin{align}
    \text{logMSE}(\mathbf{S}, \mathbf{G}) &= \text{avg}((\log_{2}\mathbf{S} - \log_{2}\mathbf{G})^2) \\
    \text{margin}(\mathbf{S}, \mathbf{G}) &= \min(\frac{\max(|\mathbf{S}-\mathbf{G}|}{\mathbf{G}}), 1)
\end{align}

where $\mathbf{S}$ and $\mathbf{G}$ represent predicted answers and golden answers respectively.
Both metrics indicate how much the calculated results deviate from the golden answers.

In Figure \ref{fig:addition_comparision}, despite failing to generate accurate answers, all three models could keep their logMSE and error margin at a very low level by adding probed $a$ and $b$, while directly accepting the output of language models would lead to results that deviate far away from the golden answers.
We think that this reveals a possibility to control the computational error of language models within a reasonable range, and will not produce results that are far too unreasonable.

We also notice that for LLaMA-2 models, adding the probed number on late layers will result in a high error margin, which may be a result of the findings in Section \ref{ssec:activation_patching}: number encoding on late layers is not used by the model.

\end{document}